\pdfoutput=1

\documentclass[11pt]{article}

\usepackage{ACL2023}

\usepackage{times}
\usepackage{latexsym}
\usepackage{pdfpages}
\usepackage{makecell}

\usepackage[T1]{fontenc}

\usepackage[utf8]{inputenc}

\usepackage{microtype}

\usepackage{inconsolata}

\usepackage[flushleft]{threeparttable}
\usepackage{adjustbox}
\usepackage{svg}
\usepackage{subcaption}

\usepackage{graphicx,float,rotating,colortbl}

\usepackage{booktabs}
\usepackage{tabularx}

\usepackage{multirow}
\usepackage{enumitem}

\usepackage{amsmath}
\usepackage{amssymb}
\usepackage{xspace}

\newcolumntype{L}[1]{>{\raggedright\let\newline\\\arraybackslash\hspace{0pt}}m{#1}}
\definecolor{sellercolor}{rgb}{0.07, 0.04, 0.56}
\definecolor{buyercolor}{rgb}{0.07, 0.04, 0.56}

\newif\ifcomments

\commentstrue

\ifcomments
    \newcommand\chenhao[1]{\textcolor{magenta}{[CT: #1]}}
    \newcommand\mourad[1]{\textcolor{red}{[MH: #1]}}
    \newcommand\rob[1]{\textcolor{violet}{[RV: #1]}}
    \newcommand\alex[1]{\textcolor{blue}{[AZ: #1]}}
\else
    \newcommand\chenhao[1]{}
    \newcommand\mourad[1]{}
    \newcommand\rob[1]{}
    \newcommand\alex[1]{}
    
\fi
\newcommand\figref[1]{Figure~\ref{#1}}
\newcommand\secref[1]{\S\ref{#1}}

\newcommand{\act}{bargaining act\xspace}
\newcommand{\actp}{bargaining acts\xspace}
\newcommand{\actc}{Bargaining act\xspace}
\newcommand{\actcp}{Bargaining acts\xspace}
\newcommand{\actt}{Bargaining Act\xspace}
\newcommand{\actnsp}{bargaining acts\xspace}
\newcommand{\actnscp}{Bargaining acts\xspace}

\title{Language of Bargaining}

\author{
Mourad Heddaya \\ University of Chicago \\  \scalebox{0.87}[0.9]{{\tt {mourad@uchicago.edu}}} \And 
Solomon Dworkin \\ University of Chicago \\  \scalebox{0.87}[0.9]{{\tt {solomon.dworkin@gmail.com}}} \And
Chenhao Tan \\ University of Chicago \\  \scalebox{0.87}[0.9]{{\tt {chenhao@uchicago.edu}}}
\AND
Rob Voigt \\ Northwestern University \\  \scalebox{0.87}[0.9]{{\tt{robvoigt@northwestern.edu}}} \And
Alexander Zentefis \\ Yale University\\  \scalebox{0.87}[0.9]{{\tt{alexander.zentefis@yale.edu}}}
}

\begin{document}
\maketitle

\begin{abstract}

Leveraging an established exercise in negotiation education, we build a novel dataset for studying how the use of language shapes bilateral bargaining.
Our dataset extends existing work in two ways: 1) we recruit participants via behavioral labs instead of crowdsourcing platforms and allow participants to negotiate through audio, enabling more naturalistic interactions; 
2) we add a control setting where participants negotiate only through alternating, written numeric offers.
Despite the two contrasting forms of communication, we find that the average agreed prices of the two treatments are identical. But 
when subjects can talk, fewer offers are exchanged, negotiations finish faster, the likelihood of reaching agreement rises, and the variance of prices at which subjects agree drops substantially.
We further propose a taxonomy of speech acts in negotiation and enrich the dataset with annotated speech acts. Our work also reveals linguistic signals that are predictive of negotiation outcomes. %

\end{abstract}

\section{Introduction}
\label{sec:introduction}

Bilateral bargaining, in the sense of a goal-oriented negotiation between two parties, is a fundamental human social behavior that takes shape in many areas of social experience.
Driven by a desire to better understand this form of interaction, a rich body of work in economics and psychology has evolved to study bargaining \citep{rubin1975social,bazerman2000negotiation,roth2020}.
However, this work has seldom paid careful attention to the use of language and its fine-grained impacts on bargaining conversations; indeed, many studies operationalize bargaining as simply the back-and-forth exchange of numerical values. 
Meanwhile, there is growing interest in bargaining in NLP oriented towards the goal of building dialogue systems capable of engaging in effective negotiation \citep{zhan2022let, fu2023improving}. 
In this work, we aim to bridge these two lines of work and develop a computational understanding of how language shapes bilateral bargaining.

To do so, building on a widely used exercise involving the bargaining over the price of a house used in negotiation education, we develop 
a controlled experimental environment to collect a dataset of bargaining conversations.\footnote{Dataset access may be requested at: \url{https://mheddaya.com/research/bargaining}} %
The treatment in our experiment is the manner in which subjects communicate:
either through alternating, written, numeric offers (the {\em alternating offers} or AO condition) or unstructured, verbal communication (the {\em natural language} or NL condition). 
Furthermore, to encourage naturalistic interactions, we recruit participants via behavioral labs and allow participants to negotiate in a conversational setting using audio on Zoom instead of crowdingsourcing text conversations as prior work has done (\citealp{asher2016discourse,lewis2017deal,he2018decoupling}). 
In total, we collect a dataset with 230 alternating-offers negotiations and 178 natural language negotiations. In contrast with \citet{he2018decoupling}'s Craigslist negotiation dataset, our natural language negotiations have an average of over 4x more turns exchanged during each conversation, so our dataset represents a richer source to explore linguistic aspects of bargaining behavior than has been presented by existing work in this area.

In addition, we enrich the dataset by annotating all the conversations with a set of negotiation-specific speech acts. 
Inspired by prior work on rhetorical strategies in negotiations \citep{chang1994speech,weigand2003b2b,twitchell2013negotiation}, 
we create a simplified taxonomy of what we term \textit{\actp} and hire undergraduate research assistants to provide annotations.
To the best of our knowledge, our dataset of speech acts in negotiations is an order of magnitude larger than existing datasets.

We first provide descriptive results based on our dataset.
Although the AO and NL conditions are conducted via different communication mechanisms, they reach the same average agreed prices. However, when subjects can talk, fewer offers are exchanged, negotiations finish faster, the likelihood of reaching agreement rises, and the variance of prices at which subjects agree drops substantially.
These observations suggest that the use of language facilitates collaboration.
We also find differences in how buyers and sellers employ \actp.

Recorded and transcribed speech provides more direct access to the intuitive attitudes and behaviors of the buyers and sellers. This enables us to identify subtle types of expression
that are predictive of negotiation outcomes and reveal underlying dynamics of negotiation. Other findings corroborate conclusions from \citet{lee2017can}, who distinguish the effectiveness of negotiators' different expressions of the same rationale.

We set up prediction tasks to predict the outcome of a negotiation based on features of the conversation and analyze the important features contributing to class differentiation.
Our results show that LIWC features provide consistently strong performance and even outperform Longformer (\citealp{beltagy2020longformer}) given the beginning of a negotiation.
Important features reveal that 
successful sellers drive and frame the conversation early on by using interrogative words to prompt buyers with targeted questions, while successful buyers convey their personal considerations and concerns while using negative expressions to push for lower prices.

In summary, we make the following contributions:
\begin{itemize}[leftmargin=*, itemsep=-2pt, topsep=0pt]
    \item We build a novel dataset of bargaining and provide annotations of \actnsp.
    \item We demonstrate that the ability to communicate using language facilitates cooperation.
    \item Our work reveals linguistic signals that are predictive of negotiation outcomes. For instance, it is advantageous to drive the negotiation, rather than to be reactive to the other party's arguments.
\end{itemize}

\section{Related Work}
\label{sec:related}

Negotiation is a growing area of study in computer science. \citet{zhan2022let} provide an excellent survey of research on negotiation dialogue systems. \citet{lewis2017deal} train recurrent neural networks to generate natural language dialogues in negotiations. \citet{he2018decoupling} propose a modular generative model based on dialogue acts.  
Our focus is on deriving computational understanding of how language shapes negotiation.

Several research disciplines have studied bilateral bargaining from different perspectives and using different tools. 
Economic theory has investigated the role of incomplete information \citep{ausubel2002bargaining} and highlighted the role of explicit communication \citep{crawford1990explicit,roth2020}.
\citet{bazerman2000negotiation} and \citet{pruitt2013negotiation} provide an overview of the psychology literature on negotiation. 
However, these studies tend to overlook the \emph{content} of the communication, with some notable exceptions \citep{swaab2011early,jeong2019communicating,lee2017can}.

The most related work to ours is \citet{lee2017can}, who study how bargaining outcomes are affected by the way a rationale is expressed. They find that expressions that hint at a constraint (e.g., ``I can't pay more'') are more effective at shaping a seller's views of the buyer's willingness to pay than critique rationales (e.g., ``it's not worth more'').

\section{Dataset}

The first contribution of our work is building the first transcript dataset of \emph{spoken} natural language bargaining between lab experiment participants. 
Our dataset extends existing datasets in four ways:
\begin{enumerate}[leftmargin=*,topsep=0pt,itemsep=0pt]
    \item Negotiation happens in spoken language, and is thus more fluid and natural, akin to real-world bargaining scenarios, such as price haggling in vendor markets, union negotiations, or diplomacy talks, while existing work is largely based on written exchanges~\citep{asher2016discourse, lewis2017deal,he2018decoupling};
    \item Our work is the first one to introduce a control condition without the use of natural language;
    \item Participants are recruited through behavioral labs at universities and their incentive structure is more high-powered (i.e., bonus earnings based on outcomes and payment exceeding the typical \$12 hourly wage) than for a crowdworker on Amazon Mechanical Turk;
    \item We supplement the transcripts with manual annotation of speech acts (see \secref{sec:negotiation_acts}).
\end{enumerate}

While contributing greatly to our understanding of negotiation, existing bargaining datasets are somewhat limited in being based on written exchanges \citep{he2018decoupling}, often in the context of a highly structured game \citep{asher2016discourse,lewis2017deal}.

\paragraph{Experiment design.} We conducted a controlled experiment whose setting reflected a common life experience: the purchase or sale of a house. 
We adapted the setting in ``Buying a House'' by Sally Blount, a popular exercise from the Dispute Resolution Research Center (DRRC) of Northwestern University's Kellogg School of Management \citep{blount2000}.\footnote{Thanks to the DRRC for kindly granting us permission to base our bargaining setting on this negotiation exercise that teaches purely distributive (i.e., zero-sum) bargaining between two parties.}
We randomly paired participants and each was assigned the role of buyer or seller. In each pairing, buyer and seller negotiated a price of the house anonymously.  
Both buyer and seller were aware of the listing price of \$240,000 and shared the same descriptions of the house and surrounding area, along with recent sales prices of comparable homes.
However, each participant was given a private valuation of the house (\$235,000 for the buyer and \$225,000 for the seller).

Participant bonus earnings depended on bargaining outcomes to incentivize subjects to engage in realistic negotiating behavior. If no agreement was reached, neither party earned bonus money. On an hourly basis, compensation seemed significant enough to influence participant behavior (i.e., at least \$40/hour was on the table per round). On average, subjects earned roughly \$23.25/hour. More details can be found in Appendix \ref{sec:app_B}. %

Each subject participated in two bargaining rounds.  In one round, a buyer-seller pair communicated via \textit{alternating offers} (AO) in an online chat that only accepted numeric entries. Each participant could choose to accept or counter each offer they received. In the other round, 
participants played the same role, either buyer or seller, but were assigned a new partner. In this round, each pair communicated in \textit{natural language} (NL) via audio only on Zoom (videos were monitored to be turned off to avoid signals from gesture and facial expressions). The subjects were restricted from disclosing their 
private value and compensation structure and informed that doing so would result in forfeiture of their earnings.\footnote{To control for the order of the two treatments affecting the bargaining outcomes, roughly half the sessions (58\% of the negotiations) first began with the round of alternating offers, whereas the other half began with the round of natural language. We did not detect any ordering effects.}
Our experiment is approved by the IRB at Yale University.

\paragraph{Preprocessing.}
We transcribed the audio from the Zoom negotiation settings using Amazon Transcribe. 
Transcription produces strictly alternating seller and buyer turns, without sentence segmentation. We use the resulting transcripts for the annotation and analyses described in this paper. We trim the end of each negotiation at the point of agreement on a final price for the house, discarding any interaction that occurs subsequently. We describe in \secref{sec:negotiation_acts} the annotation procedures that allowed us to reliably identify the point of agreement.
\begin{table}[t]
    \small
    \centering
    \begin{tabular}{lrr}
\toprule
{} &  \textbf{Alternating} & \textbf{Natural} \\
{} &  \textbf{Offers} & \textbf{Language} \\
\midrule
No. of Turns           &                     29.2 &                  42.50 \\
No. of New Offers      &                     17.9 &                   6.06 \\
No. of Repeat Offers   &                     11.3 &                   1.56 \\
Duration (min)         &                      9.5 &                   6.5 \\
Avg Turn Length (sec)  &                     28.9 &                  12.54 \\
Prob. of Agreement (\%) &                     90.0 &                  97.19 \\
Agreed Price (\$000s)   &                    229.9 &                 229.8 \\
No. of Negotiations    &                    230 &                 178 \\
No. of Unique Participants     &             460 &                 356 \\      
\bottomrule
\end{tabular}

    \caption{Descriptive Statistics Across Treatments; The table reports mean descriptive statistics of the house price negotiations in the Alternating Offer (AO) and Natural Language (NL) treatments.} 
    \label{tab:summary_stats}
\end{table}

\paragraph{Descriptive statistics.} 
Table \ref{tab:summary_stats} provides descriptive statistics of the AO and NL treatments. 
Since a failed negotiation results in no bonus for both sides, most negotiations end with a successful sale. 
Nevertheless, the probability of agreement is roughly 7 percentage points higher under NL than AO (97.2\% versus 90.0\%). A two-tailed t-test with heteroskedasticity-robust standard errors shows that the 
difference in agreement probability is significant.
Moreover, in contrast with the AO treatment, the NL treatment produces negotiations that, on average, have $\sim$1.5x more turns, but NL turns are over 50\% shorter in duration, and NL negotiations are roughly 30\% shorter in total duration and feature about 74\% fewer offers.

Surprisingly, without the ability to communicate using language, buyers and sellers are less efficient in reconciling their differences. In the AO treatment, the combination of fewer turns that are each, individually, longer in duration is telling. Interlocutors are spending more time silently strategizing and considering their next act. However, this time invested is not fruitful individually nor at the level of coordination, as exemplified by a lower probability of agreement and equivalent agreed prices among successful negotiations,
likely due to an impoverished channel of communication. 

\begin{figure}[t]	
    \begin{center}
        \includegraphics[width=1\hsize]{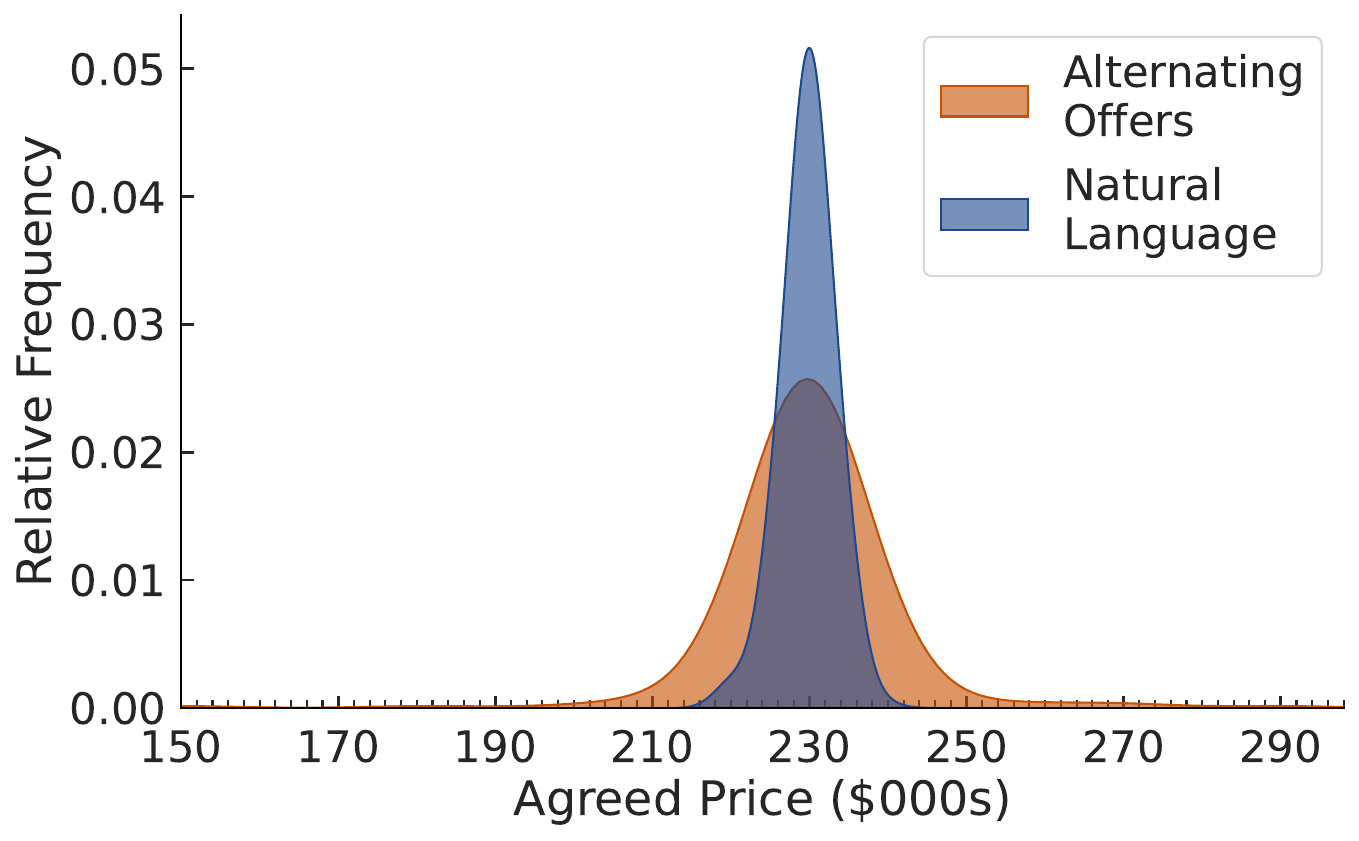}
        \caption{Gaussian kernel estimates of the distributions of agreed prices among successful negotiations.}
        \label{fig:prices_kde}
    \end{center}
\end{figure}

\begin{table*}[t]
    \small
    \centering
    \begin{tabular}{p{2cm}p{6.7cm}p{5.9cm}}
\toprule
\actt  & Definition & Example \\
\midrule
New offer & Any numerical price, not previously mentioned, that is offered by either the buyer or seller throughout the course of the negotiation. & That’s still \$30,000 out of my budget but I would be willing to pay {210,000}\\
\midrule
Repeat Offer & Any numerical price presented that is an exact repeat of a previously presented offer; in a literal sense, these are redundant offers that were already on the table. & Yeah I understand um you still think that to {240,000} is too high right \\
\midrule
Push & Any overt linguistic effort made by either party to bring the other party's offer closer to theirs. & Might just be a little too low for what I have to offer here \\
\midrule
Comparison & Evokes a difference or similarity between an aspect of the seller's house and other external houses or considerations. & Like there’s one for 213k Which is like smaller and it’s nearby so that’s closer to our budget, we’ve seen that apartment it’s not as like it’s not as furnished and it’s kind of old and so \\
\midrule
Allowance & Any time either party adjusts their offer price closer to the other party's most recent offer. An allowance may be interpreted as the accompanying interaction to a successful \textit{push} act. & I mean really like it probably should be higher than 233 but we’re willing to drop it to 233 \\
\midrule
End & End of negotiation via offer acceptance entering mutual common ground - explicitly only happens once.
& Alright 228 it is \\
\bottomrule
\end{tabular}

    \caption{\actc annotation definitions and examples.} 
    \label{tab:act_def}
\end{table*}

\figref{fig:prices_kde} 
shows that the distributions of agreed prices largely overlap between the two treatments, but the distribution in prices under NL is substantially narrower than under AO. 
Between the two treatments, the mean agreed price conditional on reaching agreement is identical (\$229.8 thousand). However, the standard deviation of agreed prices under NL is about one-third of that under AO (3.1 versus 10.4). A Fligner-Killeen (FK) \citep{fligner1976distribution} two-sample scale test shows that the standard deviation of the AO price distribution is statistically larger than the NL counterpart.

\section{\actt Annotation}
\label{sec:negotiation_acts}

Previous researchers have recognized the inherently speech-act-like character of negotiations \citep{chang1994speech,weigand2003b2b,twitchell2013negotiation}. 
Many or most utterances in a bargaining context can be thought of as taking some action with reference to the negotiation. Here we propose and present a simplified ontology of negotiation-oriented speech acts (hereafter, \textit{\actnsp}) relevant to the present context of negotiation.
Two trained undergraduate research assistants annotated all transcripts according to five \textit{\actnsp}: 1) new offer, 2) repeat offer, 3) push, 4) comparison, 5) allowance, and 6) end. Table \ref{tab:act_def} provides definitions and examples.
Note that each turn can include multiple \actp.
In addition, each speech act is also annotated with a numerical offer, if applicable.

\begin{figure}[t]	
    \begin{center}
        \includegraphics[width=1\hsize]{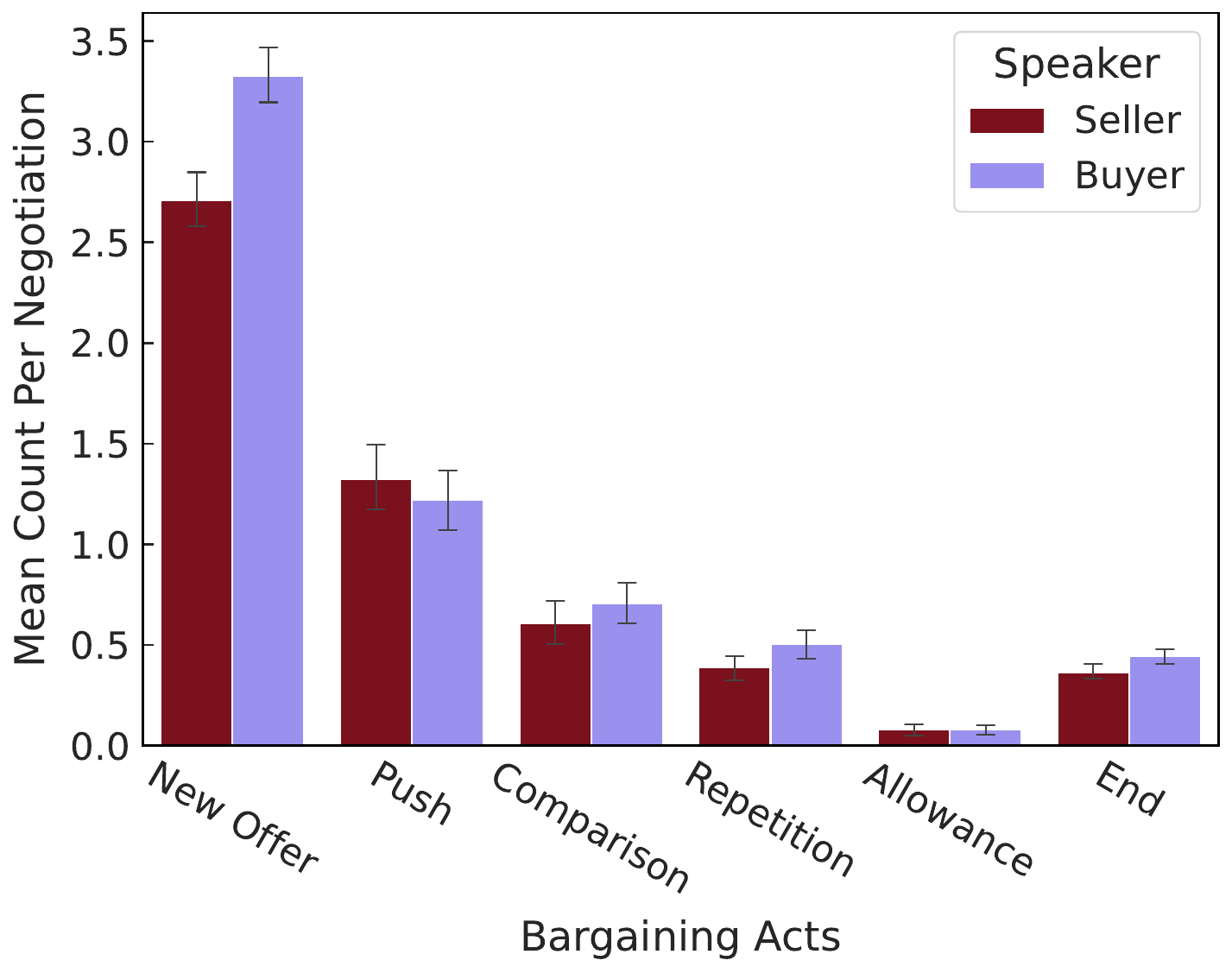}
        \caption{Distribution of Bargaining Acts. Error bars indicate standard error.}
        \label{fig:actions_distribution}
    \end{center}
   
	\medskip
\end{figure}

Twenty-four transcripts were annotated by both annotators to allow agreement to be calculated. Using MASI distance weighting (\citealp{passonneau-2006-measuring}), we found a Krippendorff's alpha (\citealp{hayes2007answering}) of 0.72, representing a high degree of agreement for a pragmatic annotation task.

\begin{figure*}[t]
     \centering
     \begin{tabular}{p{0.55\textwidth}p{0.40\textwidth}}
     \begin{subfigure}[t]{0.55\textwidth}
         \includegraphics[width=\textwidth]{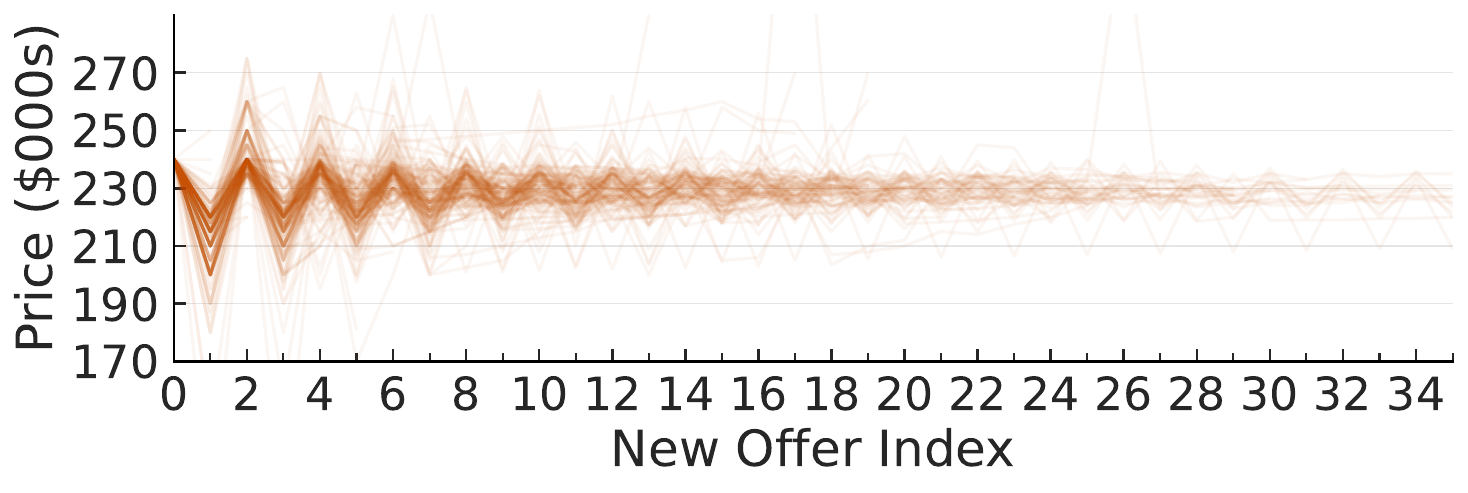}
         \caption{Alternating Offers.}
         \label{fig:ao_trajectory}    
     \end{subfigure} & \multirow{2}{*}[2cm]{
     \begin{subfigure}[t]{0.4\textwidth}\includegraphics[width=\textwidth]{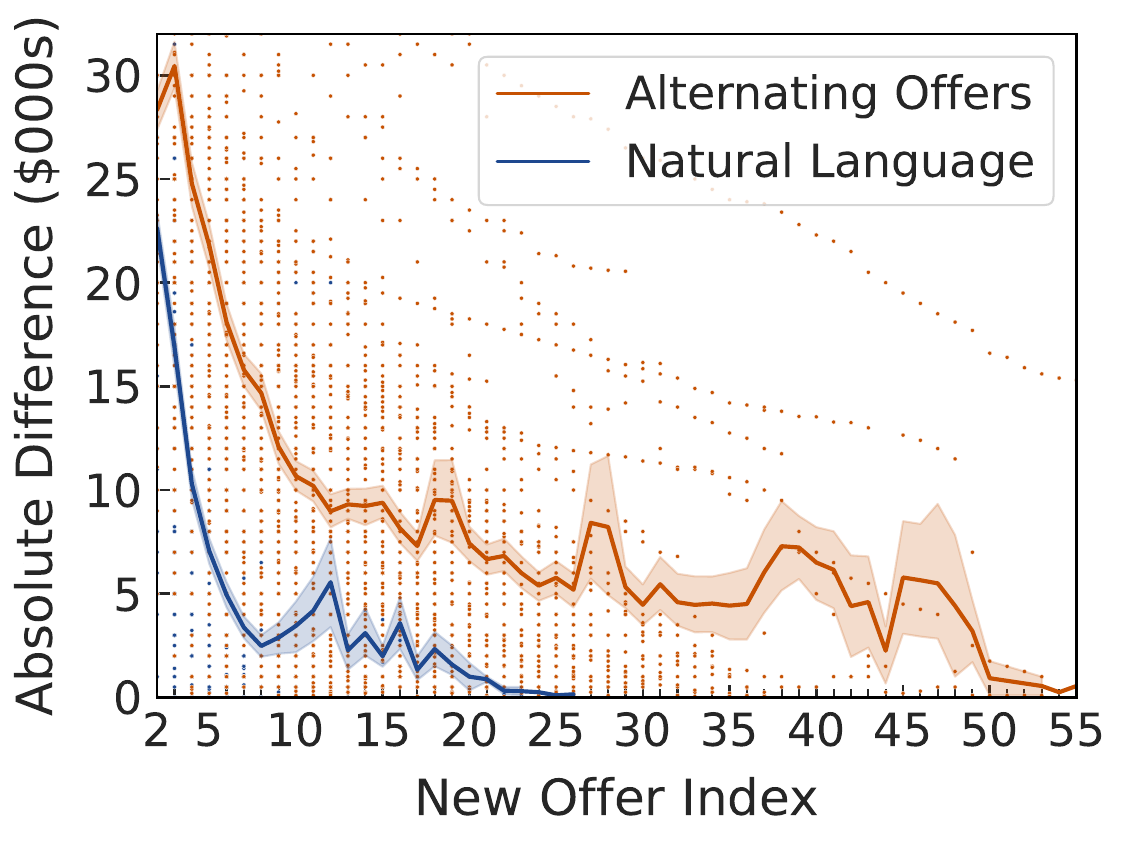}
        \caption{Absolute Differences in Consecutive New Offers.}
        \label{fig:traj_delta}
        \end{subfigure}
        }\\ 
     \begin{subfigure}[t]{0.55\textwidth}
            \includegraphics[width=\textwidth]{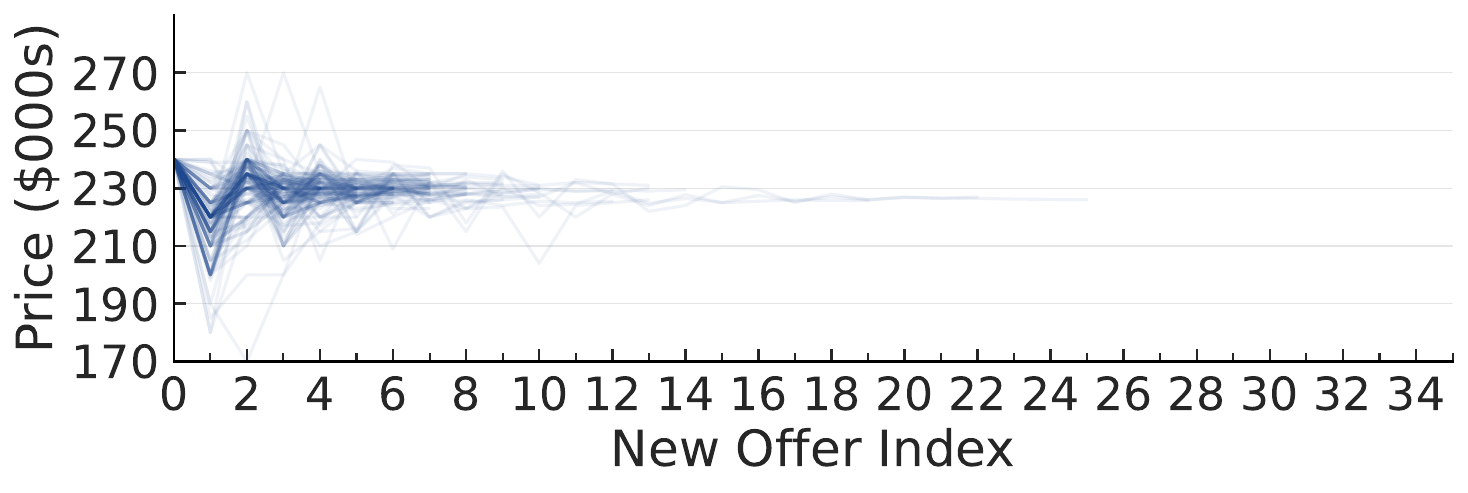}
            \caption{Natural Language.}
            \label{fig:nl_trajectory}
     \end{subfigure} & \\
     \end{tabular}
     \caption{The figure presents the trajectory of new offers in the two treatments. In \ref{fig:ao_trajectory} and \ref{fig:nl_trajectory}, each line represents a sequence of new offers exchanged between buyer and seller in a single negotiation. Only negotiations ending in agreement are included. Figure \ref{fig:traj_delta} presents the absolute differences in consecutive new offers under both treatments. Each dot represents an absolute difference in consecutive new offers within a single bargaining session. 
     }
     \label{fig:nl_ao_trajs}
\end{figure*}

\figref{fig:actions_distribution} shows that 
 \textit{new offers}, \textit{pushes}, and \textit{comparisons} are relatively more frequent and appear more consistently in all the negotiations than \textit{allowances} and \textit{repeat offers}. We note in Table~\ref{tab:summary_stats} that \textit{repeat offers} are dramatically more common in the AO condition than the NL condition (11.3 vs. 1.56 per negotiation). With linguistic context, negotiators are less likely to engage in fundamentally uncooperative behavior by simply repeating past offers over again.
 
 Comparing buyers to sellers, we observe that buyers make on average 1 more \textit{new offers} per negotiation than sellers (independent sample, heteroskedasticity robust t-test, $p=0.02$). We find no statistically significant differences between roles for the other five \actp. %

The \act annotations allow us to describe a negotiation as a sequence of offers proposed by the buyer and seller. We compare how the frequency and pattern of numerical offers differ across 1) experimental treatments (NL vs. AO) and 2) negotiation outcomes. 
We characterize different properties of the negotiations as well as their \textit{trajectories} over the course the interaction.

\figref{fig:nl_ao_trajs} reveals three general patterns on offer trajectories. First, both AO and NL bargaining feature a similar range of new offers exchanged in the early stages of the negotiation. Early on, buyers in both treatments present new offers as low as 170; and sellers, as high as 270. But extreme offers are more prevalent in AO than NL bargaining. Second, both the AO and NL trajectories exhibit a rhythmic pattern of low and high offers, which is familiar to real-world negotiations. The buyer's low offer is countered by the seller's high offer, which is then countered by the buyer's slightly increased low offer, and so on. Third, NL bargaining takes far fewer new offers to reach agreement than AO bargaining. \figref{fig:traj_delta} clearly demonstrates that NL negotiations converge quicker, with consecutive offers converging to within \$5K after 6 new offers. AO negotiations take over 40 new offer exchanges to reach a similar convergence.

\section{Predicting Negotiation Outcomes}
Finally, we set up prediction tasks to understand the relationship between the use of natural language and negotiation success. 
Overall, our models demonstrate performance gains over majority class in most settings. Surprisingly, Logistic Regression using bag-of-words and LIWC category features outperform the neural model. We observe differentiation between classification accuracy on seller only and buyer only speech, and highlight features that explain this difference.
\subsection{Experimental Setup}\label{sec:exp_setup}
\paragraph{Task.}
We consider a binary classification task with two classes: 1) ``seller win'' and 2) ``buyer win'', where a negotiation is classified by whether it concluded with an agreed price greater than \$230K or less than \$230K, respectively.
We focus on negotiations that end with an advantage for either the buyer or seller to better understand the dynamics that produce an asymmetric outcome. 
Hence, we omit the negotiations that ended with \$230K or that did not reach an agreed price.
This leaves us 119 negotiations.

As the predictive task may become trivial if we see the entire exchange, we build 10 versions of each negotiation by incrementally adding proportions of the negotiation to the input with a step size of 10\%. 
Thus, we obtain input/output pairs $(X_k, y)$ for a given negotiation, where $k=\{10\%, \dots, 100\%\}$, and each $k$ corresponds to a different prediction task; namely, whether the negotiation outcome can be predicted by the first $k$ percentage of the interaction.

\paragraph{Methods.}
We test two setups for our task. The first is a standard linear model with logistic regression. The second is an end-to-end approach using Longformer, a transformer-based model for encoding and classifying long sequences. In particular, we use the encoder and output modules of LongformerEncoderDecoder (LED) (\citealp{beltagy2020longformer}), a variant of the original Longformer model, which can encode sequences up to 16,384 tokens in length. This exceeds the maximum input length in our dataset.

In the logistic regression experiments, we treat the numerical offers as an oracle and consider three other feature sets:
1) Transcription texts;
2) \actnscp;
3) LIWC categories \cite{tausczik2010psychological}.\footnote{We also tried the union of these features, but it did not materially affect the performance.}
We represent each negotiation as a binary bag-of-words encoding of the features listed above. For \textit{\actnsp}, we construct the vocabulary based on unigrams and bigrams; 
for the other feature sets, we only include unigrams. We include bigrams for \actp to capture local combinations of \actp. %
To maintain a reasonable vocabulary size, we only consider unigrams from the transcribed text that occur in at least 5 negotiations (see Appendix \ref{sec:app_C} for total feature counts). We replace numbers mentioned in the text with a generic [NUM] token to eliminate the strongly predictive signal of new offers and focus on language instead. 
In experiments with LED, we add two special tokens [SELLER] and [BUYER] that we concatenate to the start of each turn depending on who is speaking. We make no other changes to the transcribed text. The input to LED is the concatenation of all the turns. 

\begin{figure}[t]	
    \begin{center}
        \includegraphics[width=0.4\textwidth]{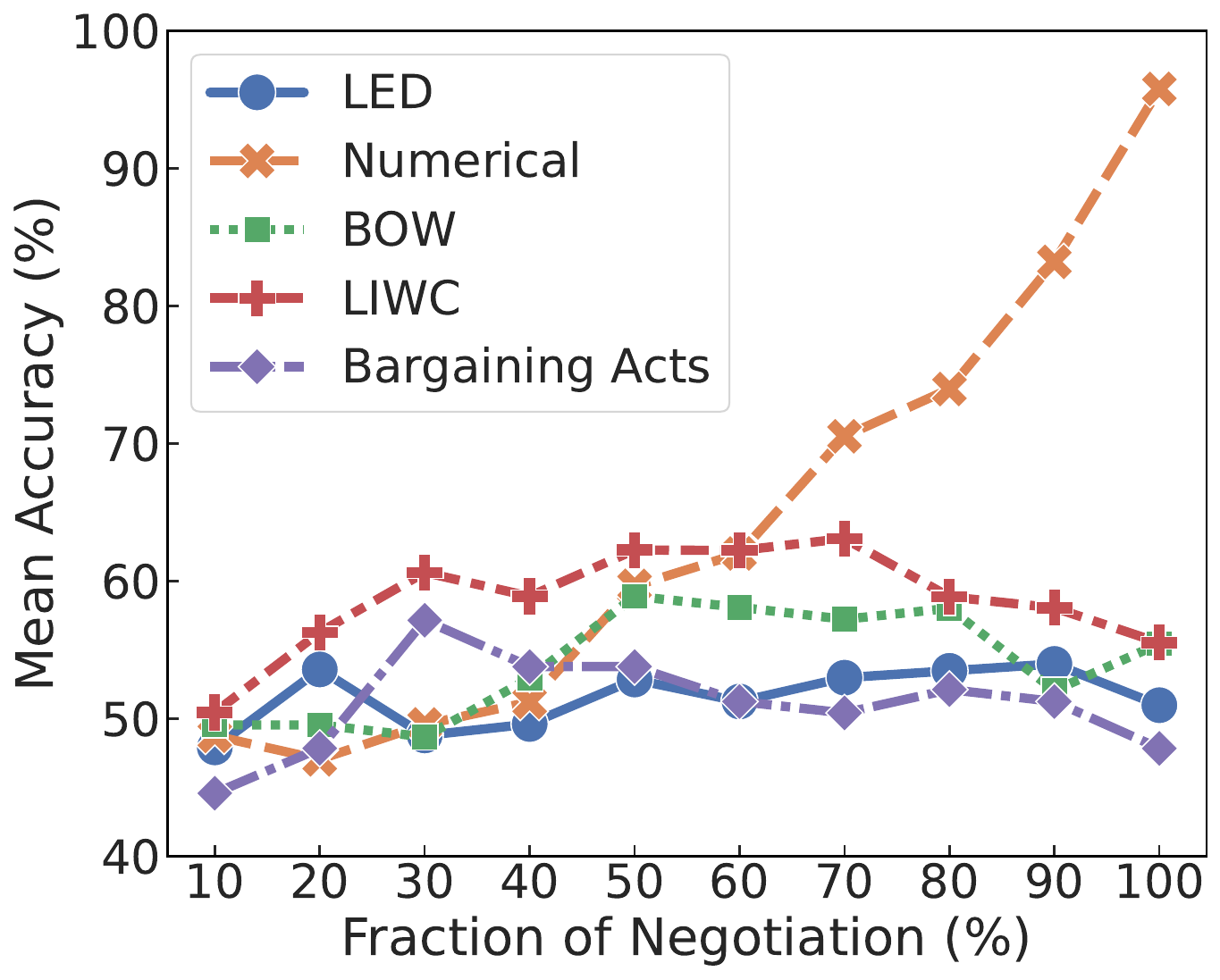}
        \caption{Overall prediction performance.}
        \label{fig:pred_all}
    \end{center}
   
	\medskip
\end{figure}

\begin{figure*}[t]
    \begin{subfigure}[t]{0.32\textwidth}    
        \centering
        \includegraphics[width=\textwidth]{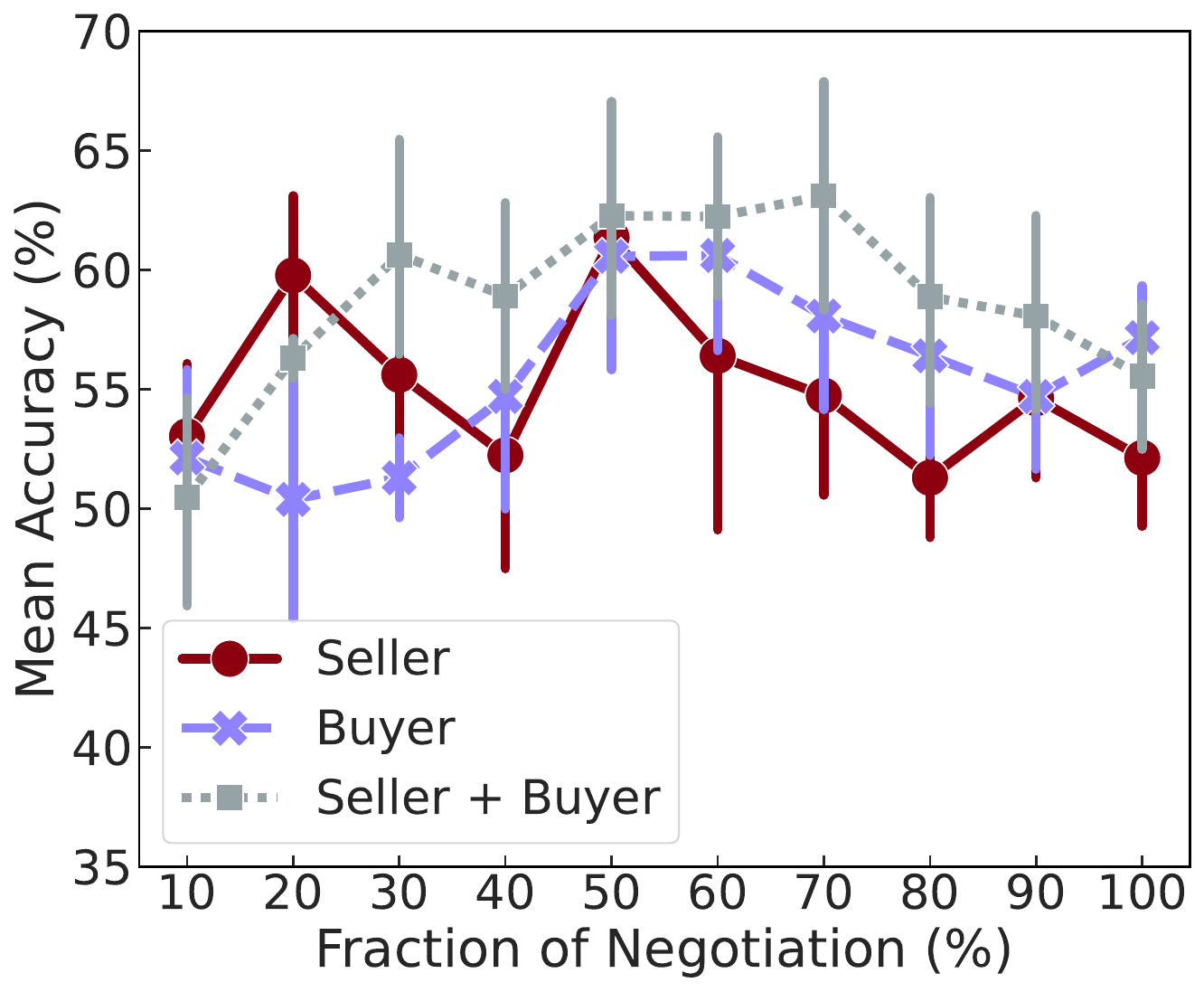}
        \caption{LIWC.
        }
        \label{fig:pred_liwc_buyer_seller}
    \end{subfigure}
    \hfill
     \begin{subfigure}[t]{0.32\textwidth}	
        \centering
        \includegraphics[width=\textwidth]{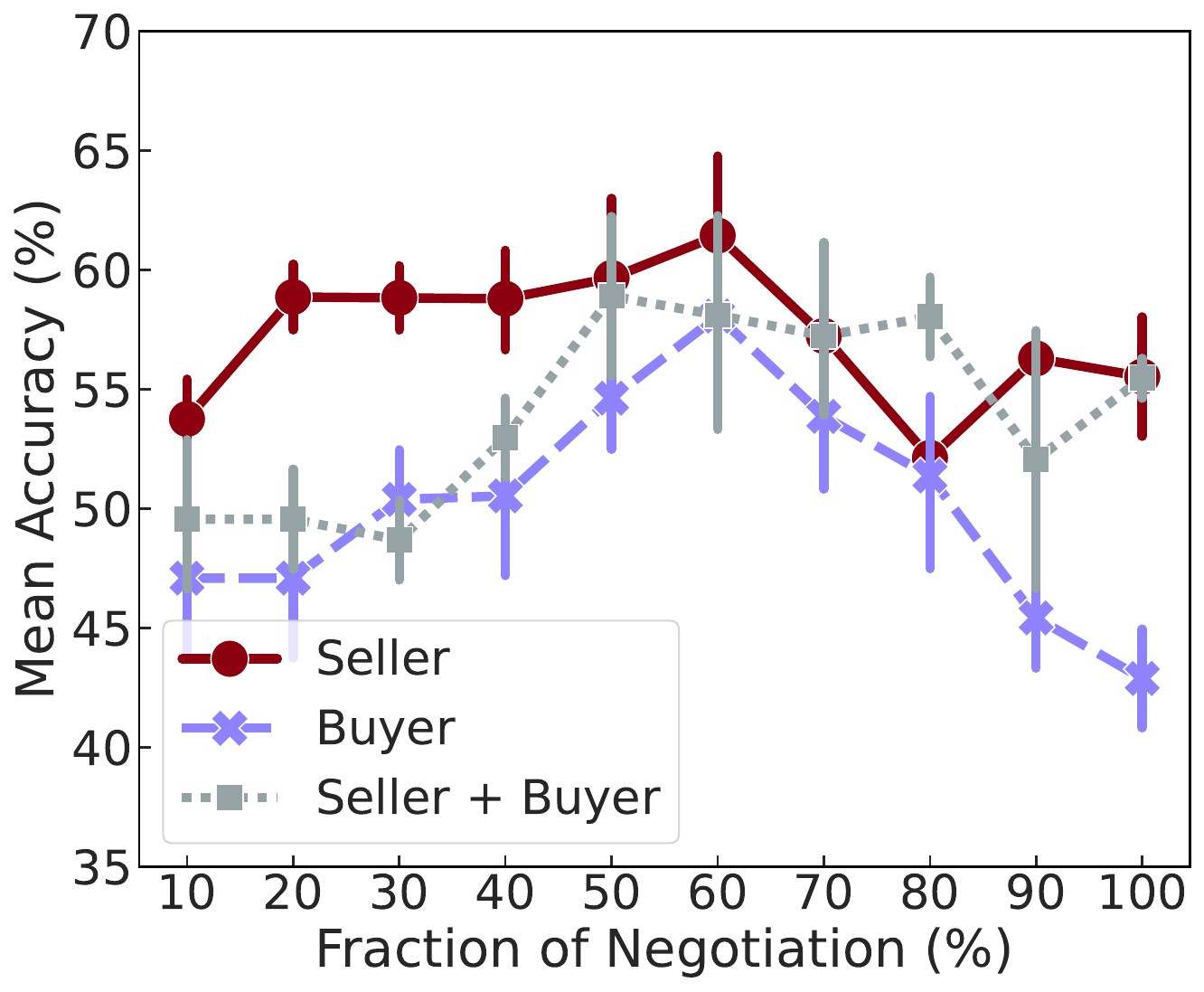}
        \caption{Transcription texts.}
        \label{fig:pred_bow_buyer_seller}
    \end{subfigure}
    \hfill
    \begin{subfigure}[t]{0.32\textwidth}    
        \centering
        \includegraphics[width=\textwidth]{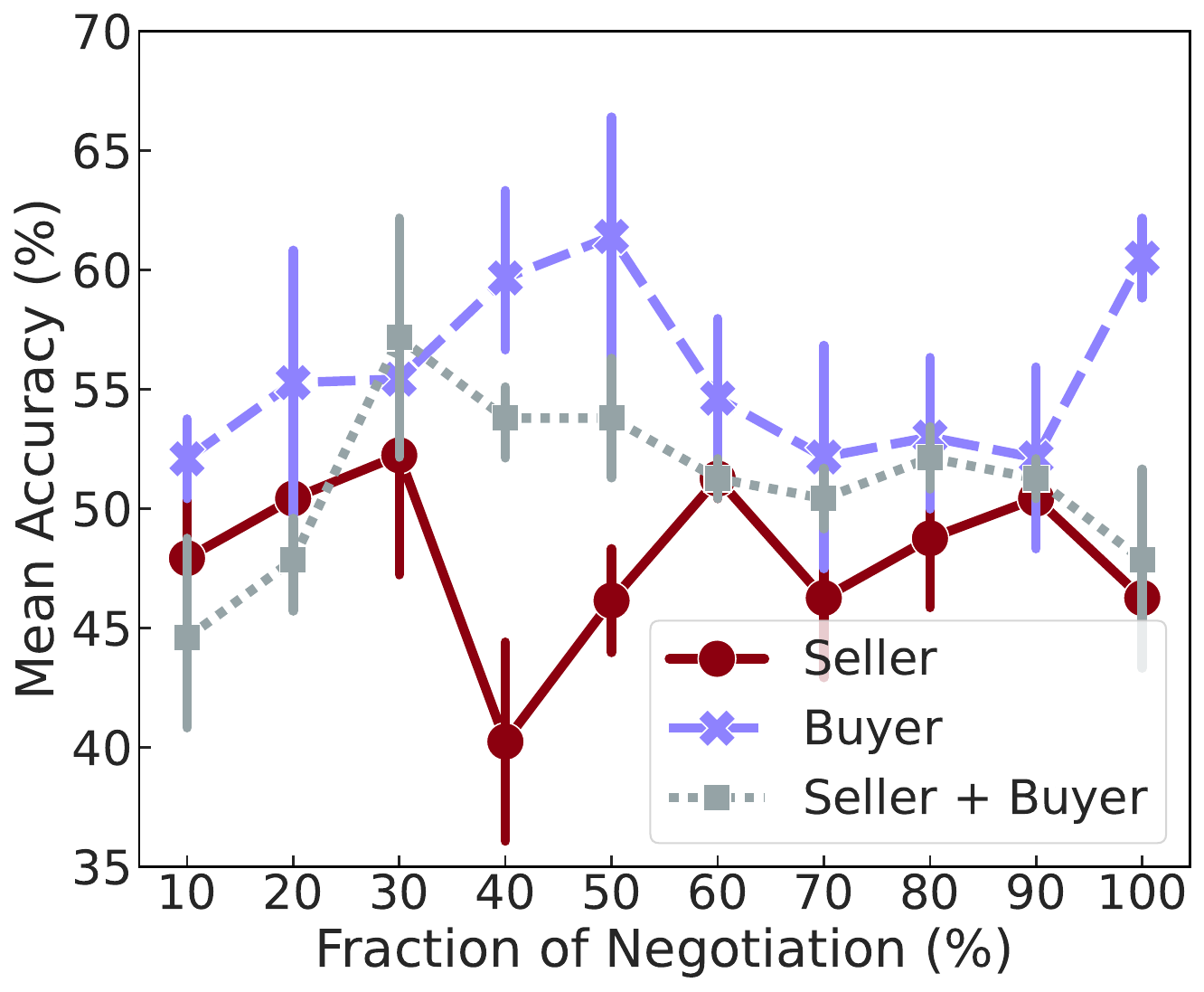}
        \caption{\actcp.}
        \label{fig:pred_acts_buyer_seller}
    \end{subfigure}   
	\caption{Buyers vs. sellers. Accuracy of Logistic Regression model across different input features, using buyer speech, seller speech, or both. Error bars indicate standard error.}
\end{figure*}

\paragraph{Evaluation.}
We use accuracy as our main evaluation metric.
In all experiments, due to the relatively small size of our dataset, we use nested five-fold cross validation for both inner and outer cross validations. For logistic regression, we grid search the best $\ell_2$ coefficient within $\{2^x\}$, where $x$ ranges over 11 values evenly spaced between –10 and 1.
We further concatenate the speaker (`buyer' or `seller') and the turn position within the negotiation. We treat these as hyper-parameters. 
We represent the position as $k$, where $k$ corresponds to a fraction of the conversation, as defined earlier. For example, the word ``house'' spoken by the seller in the first 10\% of turns in a negotiation would be tokenized as ``s1-house''.
In the LED experiments, we omit the inner cross validation and use a batch size of 4, the largest possible batch size given our memory constraints.\footnote{We use a single Nvidia A40 GPU in our LED experiments.} We select the best performing learning rate out of $\{5e-5, 3e-4, 3e-3\}$ and early stop based on training loss convergence.

\subsection{Results}
\paragraph{Predictive performance.}
We start by looking at the overall predictive performance. \figref{fig:pred_all} presents results for all models.
For the oracle condition (numerical), as expected, prediction accuracy increases monotonically and steadily as the fraction of the conversation and the corresponding numerical offers in the input increases from $10\%$ to $100\%$ of the conversation. 
As the buyer and seller converge towards an agreed price, the offers made provide strong signal about the outcome.

However, this task proves much more challenging for other models where we do not include numerical offers provided by annotators. 
One intriguing observation is that LED consistently under-performs logistic regression. 
Within logistic regression, LIWC categories outperform other features and achieve 63.1\% accuracy whereas text-based BOW features achieve a best score of 58.9\%.
Furthermore,  there is no clear trend of performance growing as the fraction of negotiation increases.
While bargaining actions under-perform other features overall, there is a notable jump in accuracy at fraction $30\%$, which we will revisit later.

\paragraph{Buyer vs. seller.}
In bilateral bargaining, an interesting question is which party drives the negotiation, and to what effect?
To further understand the role of buyer vs. seller, we only consider features of buyer texts or seller texts.

Although the performance of LIWC does not vary much for buyer and seller texts (\figref{fig:pred_liwc_buyer_seller}),
Figures \ref{fig:pred_bow_buyer_seller} and \ref{fig:pred_acts_buyer_seller} show contrasting differences in prediction accuracy for sellers and buyers at various fractions of a negotiation. Seller transcription text achieves \textasciitilde10\% higher accuracy than buyer and buyer + seller at fractions $20\%\:(p=0.01), 30\%\:(p=0.01), 90\%\:(p=0.001), 100\%\:(p=0.01)$. Meanwhile, buyer bargaining acts outperform seller acts throughout and are particularly effective at $40\%\:(p=0.008)$ and $50\%\:(p=0.03)$ of the negotiation.

\paragraph{Important features.}

To understand in greater detail which features are more helpful for prediction, we compare the fitted logistic regression models' feature coefficients.\footnote{We use the average coefficients of the five models in cross validation.} Coefficients with the largest absolute values are associated with more discriminating features. 

We first discuss features from LIWC, our best performing feature set (Table \ref{tab:lr_liwc_coefs}). 
Interrogative words spoken by the sellers at the beginning of the negotiations (``s1-interrog'') are consistently and strongly predictive of seller wins. An example use by the seller is ``so tell me about what you're looking for in a house''.
From the buyers' points of view, it appears to be disadvantageous to use informal language, such as ``mhm'', ``k'', ``yep'', and ``huh''(``b1-netspeak''), especially at the beginning of the negotiation. One interpretation could be that the buyer signals a passivity, allowing the seller to drive the conversation and establish their asking price and justification for it.
Overall, these two patterns suggest that sellers benefit from controlling the direction of the conversation early on.

\begin{table*}[ht!]
    \begin{subtable}[h]{\textwidth}
        \small
        \let\center\empty
        \let\endcenter\relax
        \centering
        \begin{tabular}[t]{>{\raggedright}p{0.17\textwidth}>{\raggedright}p{0.17\textwidth}>{\raggedright}p{0.17\textwidth}>{\raggedright}p{0.17\textwidth}>{\raggedright\arraybackslash}p{0.17\textwidth}}
\toprule
10\% &         30\% &             50\% &          70\% &             90\% \\
\midrule
\multicolumn{5}{c}{\textsc{Buyer Win}}\vspace{0.2em}\\
\textcolor[rgb]{0.549, 0.0, 0.059}{s2}-social, \textcolor[rgb]{0.549, 0.0, 0.059}{s1}-time, \textcolor[rgb]{0.549, 0.0, 0.059}{s1}-compare, \textcolor[rgb]{0.557, 0.51, 0.996}{b1}-adj, \textcolor[rgb]{0.557, 0.51, 0.996}{b1}-focuspast & \textcolor[rgb]{0.549, 0.0, 0.059}{s2}-you, \textcolor[rgb]{0.549, 0.0, 0.059}{s2}-social, \textcolor[rgb]{0.549, 0.0, 0.059}{s3}-social, \textcolor[rgb]{0.557, 0.51, 0.996}{b3}-posemo, \textcolor[rgb]{0.549, 0.0, 0.059}{s2}-space & \textcolor[rgb]{0.557, 0.51, 0.996}{b3}-posemo, \textcolor[rgb]{0.549, 0.0, 0.059}{s3}-social, \textcolor[rgb]{0.549, 0.0, 0.059}{s2}-space, \textcolor[rgb]{0.549, 0.0, 0.059}{s5}-money, \textcolor[rgb]{0.557, 0.51, 0.996}{b4}-negemo & \textcolor[rgb]{0.549, 0.0, 0.059}{s7}-home, \textcolor[rgb]{0.557, 0.51, 0.996}{b4}-negemo, \textcolor[rgb]{0.549, 0.0, 0.059}{s2}-you, \textcolor[rgb]{0.549, 0.0, 0.059}{s2}-cogproc, \textcolor[rgb]{0.557, 0.51, 0.996}{b4}-money & \textcolor[rgb]{0.557, 0.51, 0.996}{b4}-money, \textcolor[rgb]{0.549, 0.0, 0.059}{s8}-bio, \textcolor[rgb]{0.549, 0.0, 0.059}{s2}-interrog, \textcolor[rgb]{0.557, 0.51, 0.996}{b4}-negemo, \textcolor[rgb]{0.549, 0.0, 0.059}{s2}-space \\
 \midrule
 \multicolumn{5}{c}{\textsc{Seller Win}}\vspace{0.2em}\\
\textcolor[rgb]{0.557, 0.51, 0.996}{b1}-motion, \textcolor[rgb]{0.557, 0.51, 0.996}{b1}-netspeak, \textcolor[rgb]{0.557, 0.51, 0.996}{b1}-i, \textcolor[rgb]{0.557, 0.51, 0.996}{b1}-focuspresent, \textcolor[rgb]{0.549, 0.0, 0.059}{s1}-adverb & \textcolor[rgb]{0.549, 0.0, 0.059}{s1}-interrog, \textcolor[rgb]{0.557, 0.51, 0.996}{b3}-you, \textcolor[rgb]{0.557, 0.51, 0.996}{b1}-netspeak, \textcolor[rgb]{0.557, 0.51, 0.996}{b1}-motion, \textcolor[rgb]{0.557, 0.51, 0.996}{b3}-bio & \textcolor[rgb]{0.549, 0.0, 0.059}{s1}-interrog, \textcolor[rgb]{0.557, 0.51, 0.996}{b1}-netspeak, \textcolor[rgb]{0.557, 0.51, 0.996}{b3}-bio, \textcolor[rgb]{0.549, 0.0, 0.059}{s1}-you, \textcolor[rgb]{0.549, 0.0, 0.059}{s1}-conj & \textcolor[rgb]{0.557, 0.51, 0.996}{b3}-bio, \textcolor[rgb]{0.549, 0.0, 0.059}{s1}-interrog, \textcolor[rgb]{0.557, 0.51, 0.996}{b1}-netspeak, \textcolor[rgb]{0.557, 0.51, 0.996}{b3}-focusfuture, \textcolor[rgb]{0.557, 0.51, 0.996}{b3}-reward & \textcolor[rgb]{0.549, 0.0, 0.059}{s1}-interrog, \textcolor[rgb]{0.557, 0.51, 0.996}{b3}-bio, \textcolor[rgb]{0.557, 0.51, 0.996}{b1}-netspeak, \textcolor[rgb]{0.557, 0.51, 0.996}{b1}-motion, \textcolor[rgb]{0.549, 0.0, 0.059}{s4}-focuspast \\
\bottomrule
\end{tabular}

        \caption{LIWC.\vspace{1em}}
        \label{tab:lr_liwc_coefs}
    \end{subtable}
    
    \begin{subtable}[h]{\textwidth}
        \small
        \let\center\empty
        \let\endcenter\relax
        \centering
        \begin{tabular}[t]{>{\raggedright}p{0.17\textwidth}>{\raggedright}p{0.17\textwidth}>{\raggedright}p{0.17\textwidth}>{\raggedright}p{0.17\textwidth}>{\raggedright\arraybackslash}p{0.17\textwidth}}
\toprule
 10\% &         30\% &             50\% &          70\% &             90\% \\
\midrule
\multicolumn{5}{c}{\textsc{Buyer Win}}\vspace{0.2em}\\
\textcolor[rgb]{0.557, 0.51, 0.996}{b}-push, \textcolor[rgb]{0.557, 0.51, 0.996}{b}-push \textcolor[rgb]{0.557, 0.51, 0.996}{b}-new, 
  \textcolor[rgb]{0.557, 0.51, 0.996}{b}-repeat \textcolor[rgb]{0.557, 0.51, 0.996}{b}-push, \textcolor[rgb]{0.557, 0.51, 0.996}{b}-new, \textcolor[rgb]{0.557, 0.51, 0.996}{b}-new \textcolor[rgb]{0.557, 0.51, 0.996}{b}-push & \textcolor[rgb]{0.557, 0.51, 0.996}{b}-push \textcolor[rgb]{0.557, 0.51, 0.996}{b}-compare, \textcolor[rgb]{0.557, 0.51, 0.996}{b}-new \textcolor[rgb]{0.557, 0.51, 0.996}{b}-compare, \textcolor[rgb]{0.557, 0.51, 0.996}{b}-push, \textcolor[rgb]{0.557, 0.51, 0.996}{b}-compare \textcolor[rgb]{0.557, 0.51, 0.996}{b}-repeat, \textcolor[rgb]{0.557, 0.51, 0.996}{b}-push \textcolor[rgb]{0.557, 0.51, 0.996}{b}-repeat &  \textcolor[rgb]{0.557, 0.51, 0.996}{b}-new \textcolor[rgb]{0.557, 0.51, 0.996}{b}-compare, \textcolor[rgb]{0.557, 0.51, 0.996}{b}-new \textcolor[rgb]{0.557, 0.51, 0.996}{b}-repeat, \textcolor[rgb]{0.557, 0.51, 0.996}{b}-push \textcolor[rgb]{0.557, 0.51, 0.996}{b}-compare, \textcolor[rgb]{0.557, 0.51, 0.996}{b}-push, \textcolor[rgb]{0.557, 0.51, 0.996}{b}-push \textcolor[rgb]{0.557, 0.51, 0.996}{b}-new & \textcolor[rgb]{0.557, 0.51, 0.996}{b}-push \textcolor[rgb]{0.557, 0.51, 0.996}{b}-compare, \textcolor[rgb]{0.557, 0.51, 0.996}{b}-new \textcolor[rgb]{0.557, 0.51, 0.996}{b}-compare, \textcolor[rgb]{0.557, 0.51, 0.996}{b}-new \textcolor[rgb]{0.557, 0.51, 0.996}{b}-allow, \textcolor[rgb]{0.557, 0.51, 0.996}{b}-push \textcolor[rgb]{0.557, 0.51, 0.996}{b}-allow, \textcolor[rgb]{0.557, 0.51, 0.996}{b}-allow \textcolor[rgb]{0.557, 0.51, 0.996}{b}-push &             \textcolor[rgb]{0.557, 0.51, 0.996}{b}-push \textcolor[rgb]{0.557, 0.51, 0.996}{b}-compare, \textcolor[rgb]{0.557, 0.51, 0.996}{b}-push \textcolor[rgb]{0.557, 0.51, 0.996}{b}-new, \textcolor[rgb]{0.557, 0.51, 0.996}{b}-repeat \textcolor[rgb]{0.557, 0.51, 0.996}{b}-push, \textcolor[rgb]{0.557, 0.51, 0.996}{b}-push, \textcolor[rgb]{0.557, 0.51, 0.996}{b}-new \textcolor[rgb]{0.557, 0.51, 0.996}{b}-compare \\
  \midrule
  \multicolumn{5}{c}{\textsc{Seller Win}}\vspace{0.2em}\\

\textcolor[rgb]{0.557, 0.51, 0.996}{b}-new \textcolor[rgb]{0.557, 0.51, 0.996}{b}-compare, \textcolor[rgb]{0.557, 0.51, 0.996}{b}-repeat, \textcolor[rgb]{0.557, 0.51, 0.996}{b}-push \textcolor[rgb]{0.557, 0.51, 0.996}{b}-compare, \textcolor[rgb]{0.557, 0.51, 0.996}{b}-compare \textcolor[rgb]{0.557, 0.51, 0.996}{b}-push, \textcolor[rgb]{0.557, 0.51, 0.996}{b}-compare &                                          \textcolor[rgb]{0.557, 0.51, 0.996}{b}-allow \textcolor[rgb]{0.557, 0.51, 0.996}{b}-compare, \textcolor[rgb]{0.557, 0.51, 0.996}{b}-allow, \textcolor[rgb]{0.557, 0.51, 0.996}{b}-compare \textcolor[rgb]{0.557, 0.51, 0.996}{b}-allow, \textcolor[rgb]{0.557, 0.51, 0.996}{b}-compare \textcolor[rgb]{0.557, 0.51, 0.996}{b}-push, \textcolor[rgb]{0.557, 0.51, 0.996}{b}-compare & \textcolor[rgb]{0.557, 0.51, 0.996}{b}-allow \textcolor[rgb]{0.557, 0.51, 0.996}{b}-compare, \textcolor[rgb]{0.557, 0.51, 0.996}{b}-new \textcolor[rgb]{0.557, 0.51, 0.996}{b}-push, \textcolor[rgb]{0.557, 0.51, 0.996}{b}-compare \textcolor[rgb]{0.557, 0.51, 0.996}{b}-push, \textcolor[rgb]{0.557, 0.51, 0.996}{b}-repeat \textcolor[rgb]{0.557, 0.51, 0.996}{b}-new, \textcolor[rgb]{0.557, 0.51, 0.996}{b}-new &                                                                                                                                  \textcolor[rgb]{0.557, 0.51, 0.996}{b}-compare, \textcolor[rgb]{0.557, 0.51, 0.996}{b}-repeat \textcolor[rgb]{0.557, 0.51, 0.996}{b}-allow, \textcolor[rgb]{0.557, 0.51, 0.996}{b}-allow, \textcolor[rgb]{0.557, 0.51, 0.996}{b}-new, \textcolor[rgb]{0.557, 0.51, 0.996}{b}-allow \textcolor[rgb]{0.557, 0.51, 0.996}{b}-compare & \textcolor[rgb]{0.557, 0.51, 0.996}{b}-repeat \textcolor[rgb]{0.557, 0.51, 0.996}{b}-allow, \textcolor[rgb]{0.557, 0.51, 0.996}{b}-allow \textcolor[rgb]{0.557, 0.51, 0.996}{b}-compare, \textcolor[rgb]{0.557, 0.51, 0.996}{b}-compare \textcolor[rgb]{0.557, 0.51, 0.996}{b}-allow, \textcolor[rgb]{0.557, 0.51, 0.996}{b}-allow, \textcolor[rgb]{0.557, 0.51, 0.996}{b}-compare \textcolor[rgb]{0.557, 0.51, 0.996}{b}-compare \\
\bottomrule
\end{tabular}

        \caption{\actcp.}
        \label{tab:lr_acts_buyer_coefs}
    \end{subtable}
    \caption{Top features predicting negotiation success in each feature group.
     Each column corresponds to the fraction of the conversation represented in the input. Prefixes ``s-'' and ``b-'' denote seller and buyer speech, respectively. The digit in a prefix refers to the features location within the negotiation (e.g., ``b4'' refers to buyer speech in first 40\%). Top and bottom half of each table correspond to buyer and seller win features, respectively.
    }
    \label{tab:coefs_odd}
\end{table*}

Furthermore, LIWC categories ``money'', ``space'', and ``home'' are associated with buyer success. These categories consists of seller spoken words like ``area'', ``location'', ``floors'', and ``room'' and buyer spoken words like ``budget'', ``pay'', and ``priced'', among many others, which are used in reference to various aspects of the house and its price. Discussion of these subjects often revolves around the seller first justifying their asking price (``s2-space'') then the buyer disputing the houses value or their ability to afford the seller's price (``b4-money''). Additionally, buyer speech associated with negative emotions like ``unfortunately'', ``problem'', ``sorry'', ``lower'', and ``risk'' (``b4-negemo'') similarly appears 40\% into the negotiation, along with mentions of money-related words.
Buyers may benefit from moving the conversation away from concrete facts towards a discussion about what is an affordable or reasonable price for them. Crucially, successful buyers do so in a manner that portrays them as apologetic and considerate of the sellers' interests.
Given that the buyer requires movement on the asking price to succeed, they avoid language that explicitly acknowledges that the seller may be compromising their interests.
This result echoes the important role of negative expressions on negotiation outcomes by \citet{barry2008negotiator}.

\begin{table}[t]
\small
\centering
\begin{tabular}{p{0.48\textwidth}}
\toprule
{\textcolor[rgb]{0.557, 0.51, 0.996}{\textbf{Buyer:}} Okay well I really like the house but I think that The price of \$235,000 is a bit excessive especially considering um the prices of some homes that are nearby The house I'm interested in that are selling for a lot less than that Um So I would definitely want to negotiate the price Um 
\newline \textcolor[rgb]{0.549, 0.0, 0.059}{\textbf{Seller:}} Yeah How much how much was asking price again I believe it was 240 \newline \textcolor[rgb]{0.557, 0.51, 0.996}{\textbf{Buyer:}} Okay I think that a fair price would be around 218,000 Just considering other houses in the area \newline \textcolor[rgb]{0.549, 0.0, 0.059}{\textbf{Seller:}} Um But like we also have like houses newly decorated we have like two fireplaces We also have a large eat in kitchen with all the appliances And uh comparing we all the house has uh 1,846 sq ft of space and which is more than the other first listing in appendix two
}\\
\bottomrule
\end{tabular}
\caption{Example transcript excerpt.}
\label{tab:exs_c_p_1}
\end{table}

Another notable observation is that buyer-only bargaining acts are more predictive. 
To better make sense of this observation, Table~\ref{tab:lr_acts_buyer_coefs} shows important features when predicting only with buyer \act unigrams and bigrams. Most notably, \textit{new offers} and \textit{pushes} followed by \textit{comparisons} consistently appear as two of the most influential features predictive of buyer wins. 

We present an example excerpt in Table \ref{tab:exs_c_p_1} to illustrate such sequences.
In this case, the comparison is serving the role of justifying the buyer's new offer of \$218,000. 
This scenario often occurs the first time that a comparison is made by either party: It puts the seller in a position to defend their offer and provide counter-evidence in favor of dismissing the buyer's offer. 
Notably, the buyer remains clear and focused in their comparison to other comparable houses. In contrast, when the seller responds, they invoke small details to attempt to justify their original price. This defensive and overly complex response weakens their bargaining position because the relative importance of these minute details may be debated and new evidence may be introduced by the buyer to further discount the seller's position.
This conclusion complements the finding that, in contrast to the seller, the buyer is advantaged when the seller discusses details of the property, as evidenced by the LIWC feature ``s2-space''.

\paragraph{Further Evaluation. }
As an additional experiment, we train a logistic regression model on the \textsc{CraigslistBargain} dataset (\citealp{he2018decoupling}) and test it on our dataset. We include seller and buyer text, and use the same text encoding procedure described in \secref{sec:exp_setup}. In the \textsc{CraigslistBargain} dataset, the seller asking price is considered to be the seller's private value for the item being sold and the buyer's private value is separately specified. We consider the negotiation to be a seller win if the agreed price is higher than the midpoint between the two private values and a buyer win otherwise. Despite \textsc{CraigslistBargain} having a significantly larger training dataset, the maximum test accuracy across all 10 fractions of our negotiations dataset is 54\%, whereas we achieve a maximum of 60\% accuracy when we train and test on our dataset. This experiment underscores the distinctiveness of our dataset and suggests that it may contain relevant linguistic differences to other datasets within the bargaining domain.

\section{Conclusion}
In this work we design and conduct a controlled experiment for studying the language of bargaining. We collect and annotate a dataset of \textit{alternating offers} and \textit{natural language} negotiations. Our dataset contains more turns per negotiation than existing datasets and, since participants communicate orally, our setting facilitates a more natural communication environment. 
Our dataset is further enhanced with annotated \actp.
Our statistical analyses and prediction experiments confirm existing findings and reveal new insights. 
Most notably, the ability to communicate using language 
results in higher agreement rates and faster convergence.
Both sellers and buyers benefit from maintaining an active role in 
the negotiation and not being reactive to the other party.

\section*{Limitations}
We note several important limitations of this work. Perhaps most importantly, our dataset is "naturalistic," but not actually "natural" in the sense of independently occurring in the world. Though the interactions between our participants are real, the task itself is ultimately artificially constructed. In a real-world negotiation over something as valuable and significant as a house, the negotiating parties will be much more invested in the outcome than our experimental participants, whose actions change their outcome to the order of a few dollars. This difference in turn could lead real-world negotiating parties to speak differently and possibly employ substantially different strategies than we observe.

Methodologically, our study has a few limitations. Firstly our analyses are based entirely on language that has been automatically transcribed (with some manual checks), and while this helps with expense and scale, these transcripts could be missing important subtleties that influence the outcome. \citet{Koenecke2020RacialDI} uncover an important limitation of these systems, finding significant racial disparities in the quality of ASR transcriptions. The linguistic feature analysis we perform should be treated as largely exploratory, and provides suggestive and correlational rather than causal evidence for the relationship between language in the interactions and negotiation outcomes.

Lastly, there are further linguistic and interactional phenomena at play that we have not yet integrated into the analysis. For one, we have access to the audio channel of participants' actual speech, but we have not analyzed it in this work. There could very well be acoustic cues in participants' speech that are as significant to the interactions as the textual features analyzed here, particularly speech prosody which has been shown to communicate social meanings that could be highly relevant to negotiation like friendliness \cite{jurafsky2009extracting}. This particularly extends to more interactional questions of not simply who said what, but what was said in response to what and in what way. For instance, existing research has shown that acoustic entrainment in dialog (e.g., interlocutor adaptation to one another in terms of prosody) has important social associations with dialogue success \cite{levitan2012acoustic}. We leave a deeper investigation of these phenomena for future work.

\section*{Broader Impacts}
This research, collectively with prior and future related work, has the potential to advance our understanding of negotiation, a ubiquitous human activity.
Our dataset can enable future research into the dynamics of human bargaining as well as interpersonal interactions more broadly. 
By employing the findings and insights gained from such research, individuals may 
enhance their ability to negotiate effectively in various settings, such as salary negotiations, personal relationships, and community initiatives. 
Meanwhile, we must acknowledge that while a better understanding of language as an instrument in social interaction can be empowering, it may also be used as a tool for manipulation.

\section*{Acknowledgements}
We are grateful to Jessica Halten for helping us run the experiment through the Yale SOM Behavioral Lab. The experiment also would not have been possible without the excellent study session coordination by Sedzornam Bosson, Alexandra Jones, Emma Blue Kirby, Vivian Wang, Sherry Wu, and Wen Long Yang. We thank Rajat Bhatnagar for developing the web application used in the study. The human subjects experiment in this research was deemed exempt by the Yale University Institutional Review Board (IRB \#2000029151). We thank Allison Macdonald and Sammy Mustafa for their effort in the data annotation process. Their work was an invaluable contribution to the success of this research project. 
We thank all members of the Chicago Human+AI Lab and LEAP Workshop for feedback on early versions of this work. Finally, we thank all anonymous reviewers for their insightful suggestions and comments.

\bibliography{anthology,custom}
\bibliographystyle{acl_natbib}

\onecolumn
\appendix
\section*{Appendix}
\label{sec:appendix}

\section{Negotiation Excerpts}
\begin{table}[ht!]
        \tiny
        \let\center\empty
        \let\endcenter\relax
        \centering
        \resizebox{1\textwidth}{!}{\begin{tabular}{p{3cm}|p{3cm}|p{3cm}}
\toprule
{\textcolor[rgb]{0.557, 0.51, 0.996}{\textbf{Buyer:}} Okay well I really like the house but I think that The price of \$235,000 is a bit excessive especially considering um the prices of some homes that are nearby The house I'm interested in that are selling for a lot less than that Um So I would definitely want to negotiate the price Um 
\newline \textcolor[rgb]{0.549, 0.0, 0.059}{\textbf{Seller:}} Yeah How much how much Where the app was asking price again I believe it was 240 \newline \textcolor[rgb]{0.557, 0.51, 0.996}{\textbf{Buyer:}} Okay I think that a fair price would be around 218,000 Just considering other houses in the area \newline \textcolor[rgb]{0.549, 0.0, 0.059}{\textbf{Seller:}} Um But like we also have like houses newly decorated we have like two fireplaces We also have a large eat in kitchen with all the appliances And uh comparing we all the house has uh 1,846 sq ft of space and which is more than the other first listing in appendix two
}
& {\textcolor[rgb]{0.557, 0.51, 0.996}{\textbf{Buyer:}} My name is [\textit{name}] Um I am an investor looking to buy a single household family in the neighborhood Um and your house based on the information that I was given seemed like a good option And I was looking at the housing market in the area and it seems like one of the houses that closely resembles your own house has been sold for \$213,000 Um so I am interested in buying your house at a price somewhere close to that Uh price \newline \textcolor[rgb]{0.549, 0.0, 0.059}{\textbf{Seller:}} Okay perfect Um Well um That house that you're talking about was actually sold quite a while ago so the prices have appreciated quite a bit and now the asking price that we have is \$240,000 \newline \textcolor[rgb]{0.557, 0.51, 0.996}{\textbf{Buyer:}} Yeah
 }
& 
{\textcolor[rgb]{0.557, 0.51, 0.996}{\textbf{Buyer:}} I do feel like even though I agree it's a nice area it's a bit overpriced Um I mean speaking of comparisons the one I'm looking at right now listing 89 I was 6898 The selling price they're asking for is approximately 213,000 Um it has 1715 square feet And I've done the math That's a difference of 131 sq ft The difference in your asking price And my offering is to 27,000 So that equates to about \$206 per square foot Um That's the difference and I think that's a reasonable difference to make \newline \textcolor[rgb]{0.549, 0.0, 0.059}{\textbf{Seller:}} Yeah the market has been weirdly slow around here lately Um So we could come down slightly uh into the high two thirties let's say 2 39 \newline \textcolor[rgb]{0.557, 0.51, 0.996}{\textbf{Buyer:}} Um I'll raise it 214 \newline \textcolor[rgb]{0.549, 0.0, 0.059}{\textbf{Seller:}} Mhm Um Right we're gonna Stick with 239 I think } \\
\bottomrule
\end{tabular}
}
    \caption{Push following by comparison examples}
    \label{tab:exs_c_p}
\end{table}

\section{Controlled Experiment }\label{sec:app_B}
\paragraph{Compensation details summary.}
Each subject received \$10 for showing up and could earn additional bonus money per round. Bonus earnings depended on bargaining outcomes to incentivize subjects to engage in realistic negotiating behavior. 
Buyers could earn \$1 in bonus for every \$1,000 that the agreed sale price was \textit{below} the buyer's private value of \$235,000, up to a maximum of \$10 in bonus money. Sellers could earn \$1 in bonus for every \$1,000 that the agreed sale price was \textit{above} the seller's private value of \$225,000, up to a maximum of \$10. Given the private values of buyers and sellers, \$10 of surplus was available to split. No party earned bonus money in a round if an agreement was not reached. 

\section{Logistic Regression Features}\label{sec:app_C}
\begin{table}[ht!]
        \tiny
        \let\center\empty
        \let\endcenter\relax
        \centering
        \resizebox{1\textwidth}{!}{\begin{threeparttable}
\begin{tabular}{l*{11}{c}}
\toprule
{} & \multicolumn{1}{c}{Roles} & \multicolumn{1}{c}{10\%} & \multicolumn{1}{c}{20\%} & \multicolumn{1}{c}{30\%} & \multicolumn{1}{c}{40\%} & \multicolumn{1}{c}{50\%} & \multicolumn{1}{c}{60\%} & \multicolumn{1}{c}{70\%} & \multicolumn{1}{c}{80\%} & \multicolumn{1}{c}{90\%} & \multicolumn{1}{c}{100\%} \\
\midrule
\multirow{3}{*}{LIWC} & Buyer+Seller & 266 & 296 & 409 & 547 & 687 & 824 & 962 & 1105 & 1244 & 1381 \\
 & Buyer & 120 & 135 & 205 & 272 & 343 & 412 & 482 & 553 & 622 & 688 \\
 & Seller & 146 & 161 & 204 & 275 & 344 & 412 & 480 & 552 & 622 & 693 \\
 \midrule
\multirow{3}{*}{Transcription Texts} & Buyer+Seller & 261 & 589 & 1052 & 1522 & 1979 & 2420 & 2385 & 2423 & 2397 & 2375 \\
 & Buyer & 140 & 303 & 519 & 734 & 946 & 1161 & 1376 & 1554 & 1728 & 1869 \\
 & Seller & 121 & 286 & 533 & 788 & 1033 & 1293 & 1493 & 1735 & 1916 & 2116 \\
\midrule
\multirow{3}{*}{Bargaining Acts} & Buyer+Seller & 36 & 65 & 83 & 93 & 98 & 105 & 106 & 108 & 108 & 110 \\
 & Buyer & 12 & 22 & 26 & 27 & 28 & 29 & 29 & 29 & 29 & 30 \\
 & Seller &1 4 & 23 & 29 & 32 & 33 & 33 & 33 & 33 & 33 & 33 \\
\bottomrule
\end{tabular}
\end{threeparttable}
 
}
    \caption{Logistic Regression Feature Counts}
    \label{tab:lr_features}
\end{table}

\clearpage
\section{Hyperparameters}
\begin{table}[ht!]
        \tiny
        \let\center\empty
        \let\endcenter\relax
        \centering
        \resizebox{1\textwidth}{!}{\begin{threeparttable}
\begin{tabular}{l*{4}{c}}
\toprule
{Features} & \multicolumn{1}{c}{n-gram} & \multicolumn{1}{c}{Inner/Outer \textit{k}-Folds} & \multicolumn{1}{c}{Max Iterations} & \multicolumn{1}{c}{$\ell_2$ Coefficient} \\
\midrule
Numerical/BOW/LIWC  & 1 & 5 & 10k &  $\{2^x | x \in \{-10,-9, \cdots, 0, 1\}\}$   \\
Bargaing Acts  & 2 & 5 & 10k &  $\{2^x | x \in \{-10,-9, \cdots, 0, 1\}\}$   \\
\bottomrule
\\
\end{tabular}
\end{threeparttable}

}
    \caption{Logistic Regression hyperparameters. Unless otherwise specified, we use the default parameters from the Scikit-Learn LogisticRegression API.}
    \label{tab:lr_hyperparams}
\end{table}
\begin{table}[ht!]
        \tiny
        \let\center\empty
        \let\endcenter\relax
        \centering
        \resizebox{1\textwidth}{!}{\begin{threeparttable}
\begin{tabular}{l*{6}{c}}
\toprule
{Model} & \multicolumn{1}{c}{Speaker Role} & \multicolumn{1}{c}{\textit{k}-Folds} & \multicolumn{1}{c}{Max Epochs} & \multicolumn{1}{c}{Batch Size} & \multicolumn{1}{c}{Optimizer} & \multicolumn{1}{c}{Learning Rate} \\
\midrule
LED & Seller + Buyer & 5 & 20 & 4  & AdamW  & 5e-5  \\
\bottomrule
\\
\end{tabular}
\end{threeparttable}

}
    \caption{LongformerEncoderDecoder hyper-parameters. We used 3 epoch patience for early stopping based on training loss. We also implement best-practice recommendations from \citealt{zhang2021revisiting} for few-sample BERT fine-tuning. }
    \label{tab:lr_hyperparams_LED}
\end{table}
\clearpage

\clearpage
\begin{sidewaystable}
    \caption{Demographic Attributes of Study Subjects}
    \label{tab:demo_attributes}
    \let\center\empty
    \let\endcenter\relax
    \centering
    \resizebox{0.8\textwidth}{!}{\begin{threeparttable}[h]
\begin{tabularx}{\textwidth}{>{\hsize=1.5\hsize}X *{1}{>{\centering\arraybackslash\hsize=.5\hsize}X} | >{\hsize=1.5\hsize}X *{1}{>{\centering\arraybackslash\hsize=.5\hsize}X} }
\toprule
\toprule
                & Percent (\%)                      &       &  Percent (\%)                                               \\
 \cmidrule(lr){1-2}\cmidrule(lr){3-4} 
Gender          &                                           &   Employment Status                                   &       \\
\qquad Male     & 38.01                                     &   \qquad Employed, full time (40+ hrs/wk)             & 20.86 \\
\qquad Female   & 60.23                                     &   \qquad Employed, part time (up to 39 hrs/wk)        & 9.55  \\
\qquad Other    & 1.75                                      &   \qquad Unemployed, looking for work                 & 3.12  \\
Age             &                                           &   \qquad Unemployed, not looking for work             & 0.39  \\
\qquad 18-24    & 63.74                                     &   \qquad Student                                      & 65.30 \\
\qquad 25-34    & 27.49                                     &   \qquad Homemaker                                    & 0.19  \\
\qquad 35-44    & 4.09                                      &   \qquad Self-employed                                & 0.58  \\
\qquad 45-54    & 1.75                                      &   Income                                              &       \\ 
\qquad 55-64    & 2.34                                      &   \qquad \$0                                          & 14.15 \\  
\qquad 65-74    & 0.58                                      &   \qquad \$1-\$9,999                                  & 40.67 \\ 
Hispanic        & 13.45                                     &   \qquad \$10,000-\$24,999                            & 14.54 \\
Race            &                                           &   \qquad \$25,000-\$49,999                            & 13.16 \\
\qquad American Indian or Alaska Native          & 2.15     &   \qquad \$50,000-\$74,999                            & 10.22 \\
\qquad Asian    & 35.23                                     &   \qquad \$75,000-\$99,999                            & 4.52  \\
\qquad Black or African American                 & 14.29    &   \qquad \$100,000-\$149,999                          & 1.38  \\
\qquad Native Hawaiian or Other Pacific Islander & 0.20     &   \qquad \$150,000+                                   & 1.38  \\
\qquad White    & 48.73                                     &   Risk Preferences                                    &       \\
Education       &                                           &   \qquad 0 (unwilling to take risks)                  & 0.00  \\
\qquad < High School            & 0.00                      &   \qquad 1                                            & 1.36  \\
\qquad High School or GED       & 12.09                     &   \qquad 2                                            & 7.41  \\
\qquad Some college, no degree  & 36.26                     &   \qquad 3                                            & 14.62 \\
\qquad Associate degree         & 2.14                      &   \qquad 4                                            & 15.40 \\
\qquad Bachelor's degree        & 31.19                     &   \qquad 5                                            & 13.06 \\
\qquad Master's degree          & 15.20                     &   \qquad 6                                            & 15.79 \\
\qquad Doctorate or professional degree         & 3.12      &   \qquad 7                                            & 15.98 \\
Marital Status  &                                           &   \qquad 8                                            & 9.55  \\
\qquad Single (never married)   & 84.80                     &   \qquad 9                                            & 3.51  \\
\qquad Married or domestic partnership          & 13.45     &   \qquad 10 (very willing to take risks)              & 3.31  \\
\qquad Divorced                 & 1.75                      &                                                       &       \\
                                &                           &   No. of subjects w/ demographic info.                & 513   \\
                                &                           &   No. of subjects                                     & 521   \\
\bottomrule
\bottomrule
\end{tabularx}
\begin{tablenotes}[flushleft]
    \footnotesize
    \item \textit{Notes.} This table reports select demographic attributes of study subjects. Attributes were collected from a survey of subjects prior to the start of each study session. Responses were voluntary. Participants were allowed to select multiple choices for Race. All other attribute questions allowed only a single choice response. Risk preferences were elicited from the question: ``Are you generally a person who is willing to take risks or do you try to avoid taking risks?'' Respondents rated themselves on a ten-point scale from 0 (unwilling to take risks) to 10 (very willing to take risks). The percentage of respondents in each demographic category is reported, except for the number of subjects, which are the raw counts of the number of participants in the experiment across all study sessions for whom we have demographic information and the number of experiment participants in total. 
\end{tablenotes}
\end{threeparttable}

}
\end{sidewaystable}

\clearpage


\section*{Supplementary material}

# Housing Negotiation Study

## Welcome

Hello everyone, thank you for your patience as we waited for everyone to arrive. I am the study leader. You are about to participate in a study on negotiation, and you will be paid for your participation via an Amazon eGiftcard, privately emailed to you by the Yale SOM Behavioral Lab within two business days after the conclusion of the study.

Please close any program that you may have open on your computer besides Zoom. We will start with a brief instruction period.

If you have any questions during this period, please privately message the question to me, and I will answer it so that everyone can hear. In the chat, I will now send the weblink to the study instructions. Please follow along as I read the instructions.

## General

In this study, you will negotiate the price to buy or sell a house with other participants. The study consists of two rounds of negotiation. The person you negotiate with in round 1 will differ from the person you negotiate with in round 2. In each round, one of you will play the role of the house buyer, whereas the other will play the role of the house seller. In both rounds, you will play the same role as either buyer or seller.

## Compensation

I will now describe the compensation. You will receive 10 dollars for participating in the study plus have the potential to earn up to 10 dollars bonus money in each round. The amount of bonus money you earn in each round depends on the outcome of the negotiation. Your total earnings for the study are the amount that you accumulate over the two rounds. The maximum cumulative earnings are 30 dollars, whereas the minimum cumulative earnings are 10 dollars.

## Consent Form + Demographic Survey

I will now share a weblink to the consent form to participate in this study and a short demographics survey for you to complete. Please leave this Zoom session open while you complete the survey. The consent form will ask you to enter your study ID, so please be ready to enter it. If you do not consent to participating in

the study or if you are under age 18, please inform me via a private message. Once you complete the consent form and the demographics survey, please write a private message to me with the single word "done."

## ROLE PROMPTS

In a private message, I will now share information about the role of either the buyer or the seller to each of you. Take some time to read over this information, and use this information as you like in each negotiation round. As you read this information over, please keep this Zoom session open. Once you are finished reading the information, please send me a private message with the single word "done."

## ROUND 1

We will now begin the first round of the study. In this round, you will negotiate with another participant by only exchanging price offers and counter offers for the house. No other form of communication is allowed.

This negotiation will take place within a web application whose weblink I will share shortly. Upon clicking the weblink, you will begin exchanging offers with the other person. The buyer will propose the initial offer. If possible, please use the Google Chrome browser to open the weblink. While you negotiate, please keep this Zoom session open.

You and the other person will have a maximum of 15 minutes to negotiate, but you may finish before that time elapses. Once you complete the negotiation, privately message me the single word "done." If you do not reach an agreement after 15 minutes, neither of you will earn bonus money for this round.

If the role you see in the weblink differs from the one I gave you earlier, please let me know.

Please wait to begin until after I say so once all the weblinks have been sent out.

## ROUND 2

Now we will begin the second round. In this round, you will negotiate with a different person. Pairs of participants will be assigned to individual breakout rooms to negotiate. In your breakout room, you will play the same role as either the buyer or

seller as you did in the first round. But now, you and the person you are paired with will negotiate the house price by talking to each other over Zoom audio only. The conversation is not limited to the exchange of price offers. Keep your video off the entire time. The buyer should begin the negotiation.

You and the other person will have up to 15 minutes to negotiate, but you may finish before that time elapses. If you do not reach an agreement within 15 minutes, neither of you will earn bonus money for this round.

Before you start negotiating, BOTH of you should RECORD the breakout room session. To record, click record in the meeting controls at the bottom of your screen. Record to your Computer.

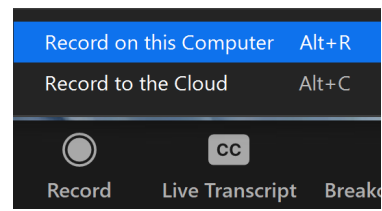

Please DO NOT start recording until after you have entered the breakout room. You may start negotiating once you hit RECORD. Once you finish negotiating, STOP the recording. Once you stop recording, leave the breakout room and return to the main room.

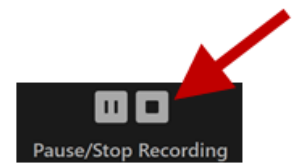

## END OF ROUND 2

The second round of negotiation is complete. Please click the weblink to a survey I will send shortly. This survey will give instructions to upload your audio recording, if you consent to do so.

Please keep this Zoom session open while you complete the survey. Once you finish the survey, privately message me the single word "done."

Before uploading your recorded audio file to the survey, please rename it "studyID.mp4" without the quotation marks, where studyID is your Study ID.

Survey to upload audio recording:
https://yalesurvey.ca1.qualtrics.com/jfe/form/SV_5opHhEOIRAigful

# Role of Buyer

(Based on "Buying a House" by Sally Blount,
Northwestern Kellogg Dispute Resolution Research Center)

**The text on this page is available only to the buyer.**

Housing values have risen rapidly in Centerville over the last few years, and you are interested in investing in a piece of real estate. Optimally, you would like to find a single family home in the $220,000 to $235,000 range, which you could rent for a few years and then resell at a profit. You recently saw an advertisement in the "Centerville Review" for a house near Centennial Park (see Appendix 1 below), which is being sold directly by its owner. Based upon the description, this house seemed like the type of investment that you are seeking. You arranged to visit the home last week. The asking price was $240,000 and you were favorably impressed.

You have since collected information on comparable houses to help you assess the worth of this house (see Appendix 2 below). You have decided that you would like to buy the house, but not at a price in excess of $235,000. In fact, you would like to buy the house at a price as close to $220,000 as possible. However, you would be willing to pay up to $235,000 before walking away from this opportunity. You will meet with the owner today to discuss buying the house.

**You cannot share the following information about your compensation with the seller. If the study coordinators learn that you have shared the following information in any form, you will forfeit your compensation.**

If you and the owner reach an agreement, you will earn $1 in bonus for every $1,000 that the agreed sale price is <u>below</u> your walk-away price of $235,000, up to a maximum of $10 in bonus money.

You will not earn any bonus money if you do not reach an agreement with the owner or if you agree to a price above $235,000.

# Appendix 1
**(available to both buyer and seller)**

**Single House**
**Listing # 90 13878**

*Description*
- 4 bedrooms + 1 recreation room + 2.5 bathrooms
- Split-level style
- Built in 1947
- 1846 square feet of space

*Inside Amenities*
- Finished hardwood floors
- 2 fireplaces
- Master bedroom with an entire wall of closets plus master bath
- Large eat-in kitchen with all appliances
- Newly decorated

*Outside Amenities*
- Comfortable & updated brick
- Beautiful landscaping
- Fenced backyard and mature trees
- Detached garage (for 2.5 cars)
- Restaurants and transportation within walking distance
- Near Hastings & Centennial parks

**Asking Price: $240,000**

# Appendix 2
**(available to both buyer and seller)**

**Prices of neighboring homes with similar characteristics**

| Listing # | Selling Price | Square Feet |
|---|---|---|
| 89 06898 | $213,300 | 1715 |
| 89 04725 | $233,600 | 1875 |
| 89 08614 | $239,600 | 1920 |

# Role of Seller

(Based on "Buying a House" by Sally Blount,
Northwestern Kellogg Dispute Resolution Research Center)

**The text on this page is available only to the seller.**

You have owned your house near Centennial Park in Centerville for several years (see Appendix l below). You originally purchased it for $155,000. To save on commissions, you have decided to sell the house yourself. After discussions with your friends who are real estate investors, you have set an asking price of $240,000. The house has been on the market for one month, and you have not yet had a firm offer.

You have always believed that you have one of the nicest houses in the Centennial Park area. You also think your house is favorably priced in comparison to comparable homes in Centerville. It has been several months since the last house was sold in the Centennial Park area. Thus, your asking price on a per square foot basis is higher (see Appendix 2 below).

Since your house has been on the market for several weeks, you have decided that you would settle for any offer that yielded at least $225,000. However, you would prefer to sell as close to $240,000 as possible. You would rather hold on to the house than sell below $225,000. Last week a prospective buyer visited your home and showed a keen interest in buying the house. You will meet with that prospective buyer today to discuss selling the house.

**You cannot share the following information about your compensation with the buyer. If the study coordinators learn that you have shared the following information in any form, you will forfeit your compensation.**

If you and the buyer reach an agreement, you will earn $1 in bonus for every $1,000 that the agreed sale price is <u>above</u> your minimum sale price of $225,000, up to a maximum of $10 in bonus money.

You will not earn any bonus money if you do not reach an agreement with the buyer or if you agree to a price below $225,000.

# Appendix 1
**(available to both buyer and seller)**

**Single House**
**Listing # 90 13878**

*Description*
- 4 bedrooms + 1 recreation room + 2.5 bathrooms
- Split-level style
- Built in 1947
- 1846 square feet of space

*Inside Amenities*
- Finished hardwood floors
- 2 fireplaces
- Master bedroom with an entire wall of closets plus master bath
- Large eat-in kitchen with all appliances
- Newly decorated

*Outside Amenities*
- Comfortable & updated brick
- Beautiful landscaping
- Fenced backyard and mature trees
- Detached garage (for 2.5 cars)
- Restaurants and transportation within walking distance
- Near Hastings & Centennial parks

**Asking Price: $240,000**


# Appendix 2
**(available to both buyer and seller)**

**Prices of neighboring homes with similar characteristics**

| Listing # | Selling Price | Square Feet |
|---|---|---|
| 89 06898 | $213,300 | 1715 |
| 89 04725 | $233,600 | 1875 |
| 89 08614 | $239,600 | 1920 |

**Welcome to the negotiation bidding page! You are the buyer. The negotiation process starts after you propose a price.**

**Role of Buyer**

(Based on "Buying a House" by Sally Blount, Northwestern Kellogg Dispute Resolution Research Center)

Housing values have risen rapidly in Centerville over the last few years, and you are interested in investing in a piece of real estate. Optimally, you would like to find a single family home in the $220,000 to $235,000 range, which you could rent for a few years and then resell at a profit. You recently saw an advertisement in the "Centerville Review" for a house near Centennial Park (see Appendix 1), which is being sold directly by its owner. Based upon the description, this house seemed like the type of investment that you are seeking. You arranged to visit the home last week. The asking price was $240,000 and you were favorably impressed.

You have since collected information on comparable houses to help you assess the worth of this house (see Appendix 2). You have decided that you would like to buy the house, but not at a price in excess of $235,000. In fact, you would like to buy the house at a price as close to $220,000 as possible. However, you would be willing to pay up to $235,000 before walking away from this opportunity. You will meet with the owner today to discuss buying the house.

**You cannot share the following information with the seller. If the study coordinators learn that you have shared the following information in any form, you will forfeit your compensation.**

If you and the owner reach an agreement, you will earn $1 in bonus for every $1,000 that the agreed sale price is below your walk-away price of $235,000, up to a maximum of $10 in bonus money.

You will not earn any bonus money if you do not reach an agreement with the owner or if you agree to a price above $235,000.

Time Remaining: 15:00

Type your price (Enter a number)   Send

Figure 6: Buyer web app page.

**Welcome to the negotiation bidding page! You are the seller. The negotiation process starts after the buyer proposes a price.**

**Role of Seller**

(Based on "Buying a House" by Sally Blount, Northwestern Kellogg Dispute Resolution Research Center)

You have owned your house near Centennial Park in Centerville for several years (see Appendix 1). You originally purchased it for $155,000. To save on commissions, you have decided to sell the house yourself. After discussions with your friends who are real estate investors, you have set an asking price of $240,000. The house has been on the market for one month, and you have not yet had a firm offer.

You have always believed that you have one of the nicest houses in the Centennial Park area. You also think your house is favorably priced in comparison to comparable homes in Centerville. It has been several months since the last house was sold in the Centennial Park area. Thus, your asking price on a per square foot basis is higher (see Appendix 2).

Since your house has been on the market for several weeks, you have decided that you would settle for any offer that yielded at least $225,000. However, you would prefer to sell as close to $240,000 as possible. You would rather hold on to the house than sell below $225,000. Last week a prospective buyer visited your home and showed a keen interest in buying the house. You will meet with that prospective buyer today to discuss selling the house.

**You cannot share the following information with the buyer. If the study coordinators learn that you have shared the following information in any form, you will forfeit your compensation.**

If you and the buyer reach an agreement, you will earn $1 in bonus for every $1,000 that the agreed sale price is above your minimum sale price of $225,000, up to a maximum of $10 in bonus money.

You will not earn any bonus money if you do not reach an agreement with the buyer or if you agree to a price below $225,000.

Time Remaining: 15:00

Waiting for other person's response

Figure 7: Seller web app page.


\begin{figure}[t]   
   \begin{center}
       \includegraphics[width=1\hsize]{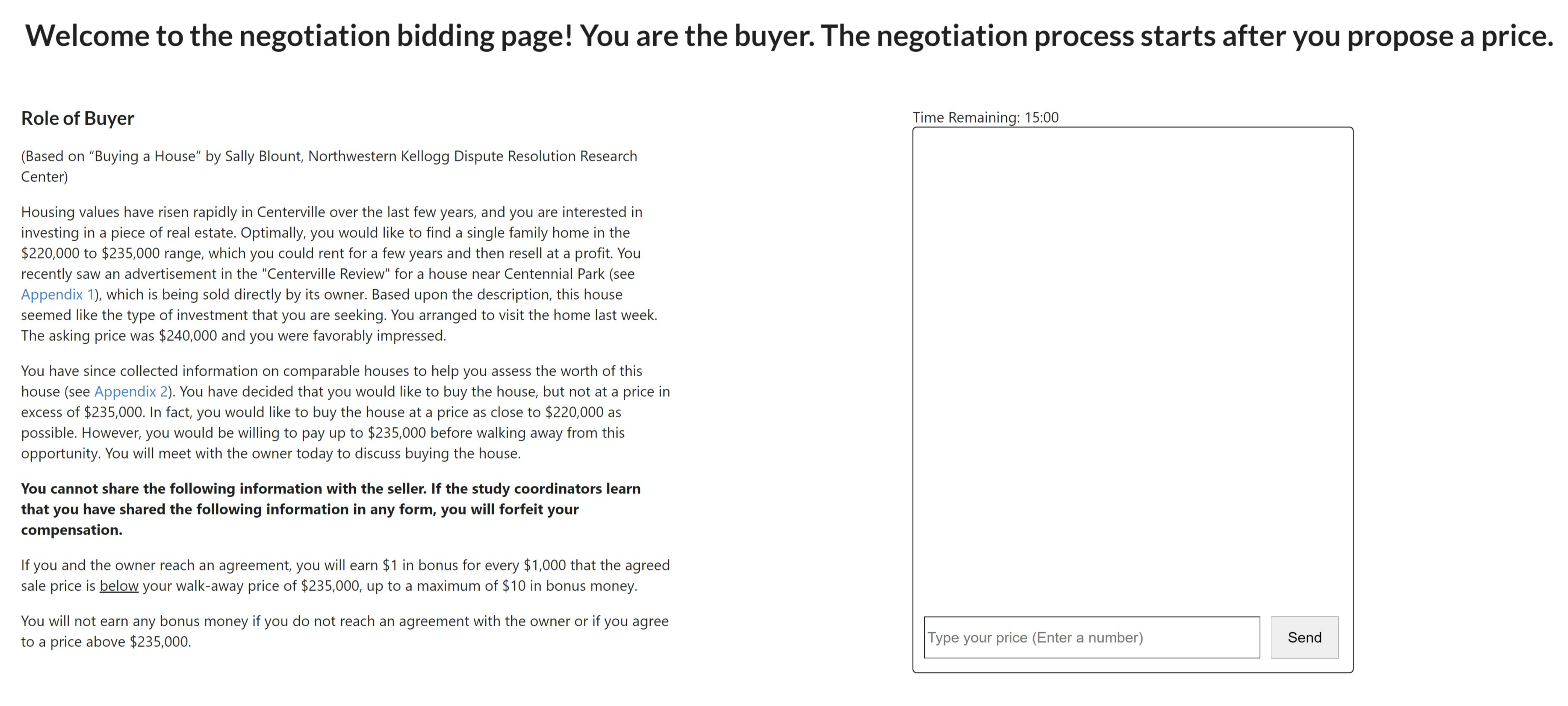}
       \caption{Buyer web app page.}
       \label{fig:buyer_web_app}
   \end{center}
\end{figure}

\begin{figure}[t]   
   \begin{center}
       \includegraphics[width=1\hsize]{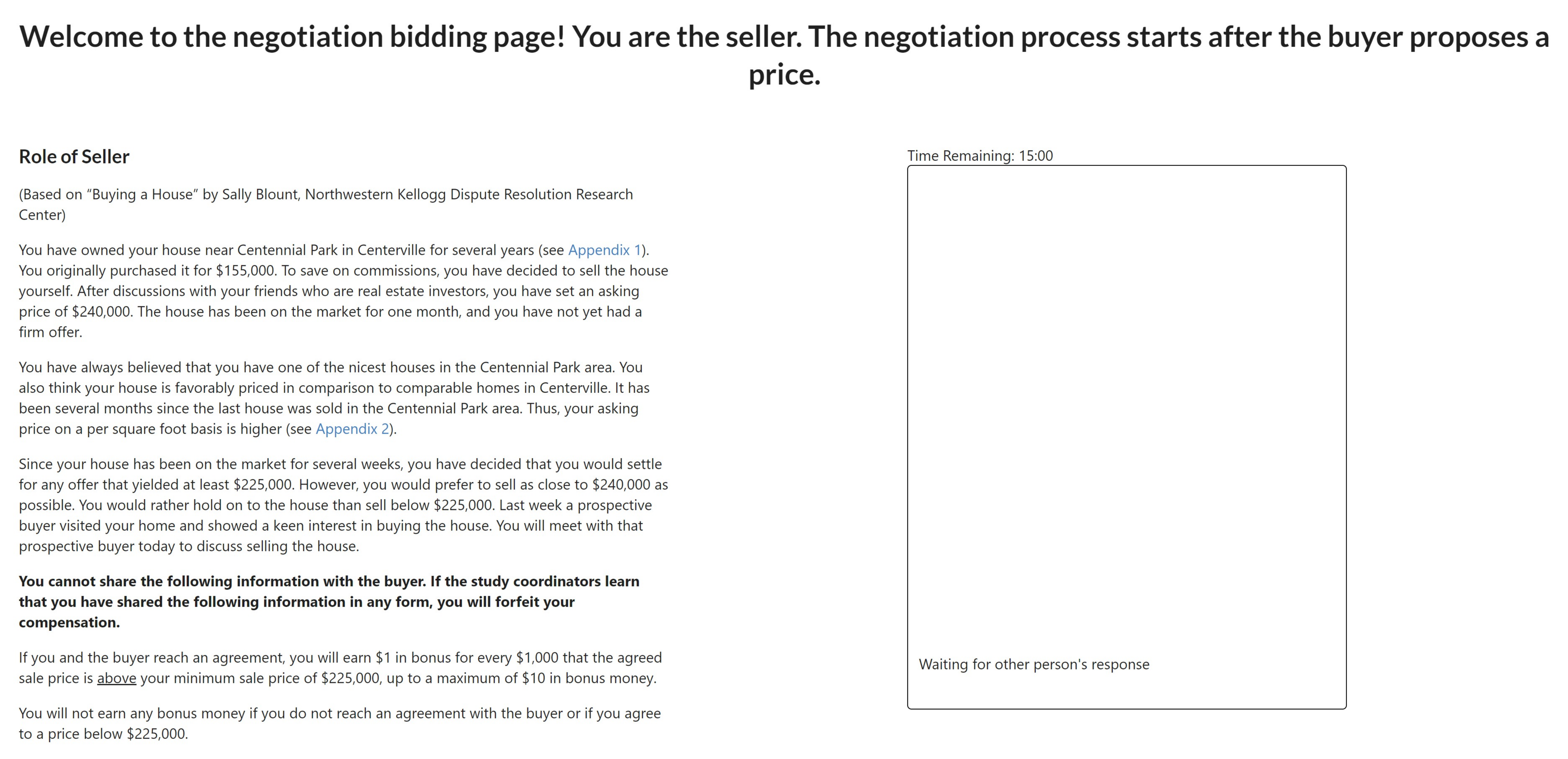}
       \caption{Seller web app page.}
       \label{fig:seller_web_app}
   \end{center}
\end{figure}

\end{document}